\documentclass[a4paper,conference]{IEEEtran}
\ifCLASSINFOpdf
\else
\fi
\usepackage[ruled,vlined]{algorithm2e}
\usepackage{times}
\usepackage{epsfig}
\usepackage{graphicx}
\usepackage{amsmath}
\usepackage{amssymb}
\usepackage{xcolor}
\usepackage{amsfonts}
\usepackage{booktabs} 
\usepackage{tabu}
\usepackage{verbatim}
\usepackage{subfig}
\usepackage{caption}
\usepackage{array, multirow}
\usepackage{makecell}
\usepackage{easylist}
\usepackage{xcolor}

\usepackage[pagebackref=true,breaklinks=true,letterpaper=true,colorlinks,bookmarks=false]{hyperref}

\pdfoutput=1

\DeclareMathOperator\arctanh{arctanh}

\DeclareMathOperator*{\argmax}{argmax}

\DeclareMathOperator{\sign}{sign}

\DeclareMathOperator{\KL}{D_{\textsc{kl}}}

\begin{document}
\title{Active Sampling for Pairwise Comparisons via Approximate Message Passing and Information Gain Maximization}

\author{\IEEEauthorblockN{Aliaksei Mikhailiuk\IEEEauthorrefmark{1}, Clifford Wilmot\IEEEauthorrefmark{1}, Maria Perez-Ortiz\IEEEauthorrefmark{3}\IEEEauthorrefmark{1}, Dingcheng Yue\IEEEauthorrefmark{1}, Rafa{\l} K. Mantiuk\IEEEauthorrefmark{1}}
\IEEEauthorblockA{\IEEEauthorrefmark{1}Department of Computer Science University of Cambridge
Cambridge, United Kingdom
\\
Email: am2442@cam.ac.uk, cliffw29@outlook.com, \{dy276, rafal.mantiuk\}@cam.ac.uk}
%\IEEEauthorblockA{ \IEEEauthorrefmark{2}Unaffiliated, 
%London, United Kingdom,
%email: cliffw29@outlook.com
%}
\IEEEauthorblockA{\IEEEauthorrefmark{3}Department of Computer Science, University College London, 
London, United Kingdom\\
Email: maria.perez@ucl.ac.uk}

}

%Telephone: (800) 555--1212, Fax: (888) 555--1212}


\maketitle

\begin{abstract}
Pairwise comparison data arise in many domains with subjective assessment experiments, for example in image and video quality assessment. In these experiments observers are asked to express a preference between two conditions. However, many pairwise comparison protocols require a large number of comparisons to infer accurate scores, which may be unfeasible when each comparison is time-consuming (e.g. videos) or expensive (e.g. medical imaging). This motivates the use of an active sampling algorithm that chooses only the most informative pairs for comparison. In this paper we propose ASAP, an active sampling algorithm based on approximate message passing and expected information gain maximization. Unlike most existing methods, which rely on partial updates of the posterior distribution, we are able to perform full updates and therefore much improve the accuracy of the inferred scores. The algorithm relies on three techniques for reducing computational cost: inference based on approximate message passing, selective evaluations of the information gain, and selecting pairs in a batch that forms a minimum spanning tree of the inverse of information gain. We demonstrate, with real and synthetic data, that ASAP offers the highest accuracy of inferred scores compared to the existing methods. We also provide an open-source GPU implementation of ASAP for large-scale experiments.% that outperforms the state-of-the-art methods both in terms of active sampling accuracy and computation time.
\end{abstract}


% \fi
\IEEEpeerreviewmaketitle

\section{Introduction}
The fields of subjective assessment and preference aggregation are concerned with measuring and modeling human judgments. Participants usually \emph{rate} a set of stimuli or conditions according to some criteria, or \emph{rank} a subset of them. Rating is inherently more complex for participants than ranking \cite{Tsukida2011,Shah2015,Mantiuk2012a,Ye2014}. Thus, comparative judgment experiments are gaining attention in subjective assessment and crowd-sourced experiments, e.g. for image quality assessement.
The simplest form of ranking experiments is comparing conditions in pairs (pairwise comparison protocol), and hence it is the most common ranking choice. Here observers are asked to choose one out of two conditions according to some criteria. As opposed to rating, in which conditions are mapped directly to a scale by computing mean opinion scores, we need to model and infer the latent scores from pairwise comparisons. This problem is known as psychometric scaling. Models used for scaling typically rely on the assumptions of Thurstone's model~\cite{Thurstone1927} or Bradley-Terry's  model~\cite{Bradley1952}. The main limitation of pairwise comparison experiments is that for $n$ conditions there are $\binom{n}{2}=n(n-1)/2$ possible pairs to compare, which makes collecting all comparisons too costly for large $n$. However, \emph{active sampling} can be used to select the most informative comparisons, minimizing experimental effort while maintaining accurate results.

The need for an efficient active sampling algorithm for preference aggregation is motivated by the recent spread of applications reliant on: i) user preferences (i.e. recommendation systems, information retrieval and relevance estimation) \cite{7023415}; ii) matchmaking in gaming systems such as TrueSkill for Xbox Live \cite{NIPS2006_3079} and Elo for chess and tennis tournaments \cite{doi:10.1111/1467-9876.00159}; iii) psychometric experiments for behavioural psychology \cite{doi:10.1348/000711003321645412} and iv) quality of experience (e.g. image and video quality) \cite{Prashnani_2018_CVPR,8578166,Ponomarenko2015,Ye2014}.

State-of-the-art active sampling methods are typically based on information gain maximization \cite{AAAI124747, GLICKMAN2005279, crowdbt2013, Ye2014, HybridMSTRAFAL, Xu2018HodgeRankWI}, where pairs in each trial are selected to maximize the weighted change of the posterior distribution of the scale. %where the information obtained from the comparison is maximized with respect to the accuracy of the scale.\RM{where the pairs are selected to maximize the change of the posterior distribution in each trial.}
However, these are computationally expensive for a large number of conditions ($n$), as they require computing the posterior distribution for $n(n-1)/2$ pairs at every iteration of the algorithm. To make active sampling computationally feasible, most existing techniques update the posterior distribution only for the pairs that were selected for the next comparison. We show that this leads to a sub-optimal choice of pairs and worse accuracy as the number of measurements increases. To address this problem, we substantially reduce the computational cost of active sampling by using approximate message passing for inference, and by computing the expected information gain only for the subset of the most informative pairs. The reduced computational overhead allows us to update the full posterior distribution at every iteration, thus greatly improving the accuracy. To ensure balanced design and allow for a batch sampling mode, we sample the pairs from a minimum spanning tree as in \cite{HybridMSTRAFAL}. The proposed technique (ASAP - Active SAmpling for Pairwise comparisons) results in the most accurate psychometric scale, especially for a large number of measurements. Moreover, the algorithm has a structure that is easy to parallelize, allowing for a fast GPU implementation. We show the benefit of using full posterior update by comparing to an approximate version of the algorithm (ASAP-approx), that, similar to other methods relies on online posterior update. 
Our main contributions are: A) an analysis of existing active sampling methods for pairwise comparison experiments under a range of condition score distributions, using both synthetic and real image and video quality assessment data; B) a novel active sampling method (ASAP), offering the highest accuracy of the scale; and C)  along with the paper we include an implementation of 9 algorithms\footnote{\url{https://github.com/gfxdisp/asap}}, providing an open-source software for active sampling in pairwise comparison experiments and including the first GPU implementation of such a method.

\section{Related work}

Comparative judgment experiments arise in ranking (ordering conditions) and scaling applications (putting conditions on a scale where distances convey meaning). Suppose we aim to compare a set of $n$ conditions $S = \{o_1,\ldots,o_n\}$ (conditions being images, players, etc.) that are evaluated according to a feature or characteristic (subjective measurements such as aesthetics, relevance, quality, etc.) with unknown underlying ground truth scores $\mathbf{s}=(s_1,\ldots,s_n)$, $s_i \in \mathbb{R}$. In this paper, we simply refer to these as quality scores. The simplest experimental protocol is to compare pairs $(o_i,o_j)$, $o_i, o_j \in S$, $i \neq j$ (referred to as pairwise comparisons).
Although other works exist, e.g.\ estimating total or partial order \cite{heckel2018approximate,heckel2019,jamieson2011,yue2011,szorenyi2015}, this paper is focused on active sampling for psychometric scale construction, which uses pairwise comparisons to estimate quality scores $\hat{\mathbf{s}}$ that approximate $\mathbf{s}$. This section discusses related work, divided into four groups, based on the type of approach: passive, sorting, information-gain and matchmaking. The methods tested in the experiments are highlighted in bold face. We also distinguish between \emph{sequential} methods \textemdash where the next pair is generated only upon receiving the outcome for the preceding pair \textemdash and batch, or parallel methods \textemdash where a batch of comparison pairs is generated and outcomes can be obtained in parallel. Batch methods are preferred in crowd-sourcing, where multiple conditions are distributed to participants in parallel.

\paragraph{Passive approaches}
When every condition is compared to every other condition the same number of times, the experimental design is referred to as \textit{full pairwise comparisons} ({FPC}). Such an approach is impractical, as it requires $n(n-1)/2$ comparisons per participant.
Another approach, \textit{nearest conditions} ({NC}), relies on the idea that conditions that are similar in quality are more informative for constructing the quality scale \cite{zerman:hal-01654133}. Thus, if the approximate ranking is known in advance, one can compare only the conditions that are neighbours in the ranking. Such initial ranking, however, may not be available in practice. 

\paragraph{Sorting approaches} Similar to NC, sorting-based methods rank the conditions, then compare those that are of similar quality. Authors in \cite{doi:10.1117/1.1344187} proposed an active sampling algorithm using a binary tree. Every new condition descends down the tree, branching depending on whether it is better or worse than the condition in the current node. Authors in \cite{Maystre2017} applied \textbf{Quicksort} \cite{10.1093/comjnl/5.1.10} using pairwise comparisons as the comparison operator. 

Recently, \cite{Ponomarenko2015} used the \textbf{Swiss system} in chess to rank subjective assessment of visual quality.  The Swiss system first chooses random conditions to compare, then sorts the conditions to find pairs that are similar. 
%Sorting variants were proposed for chess, e.g. \textbf{Swiss system}, to match players of similar skill, but were also used for subjective assessment experiments \cite{Ponomarenko2015}. The first comparisons are chosen at random; then, players are sorted and those of similar skill compete in pairs. 
A related method is the Adaptive Rectangular Design (ARD) \cite{P915} which allows comparison of conditions far apart on the quality scale in later stages of an experiment. The work of \cite{chen2016} takes a different approach, where active sampling (\textbf{AKG}) is based on the Bayesian decision process maximising Kendall's tau rank correlation coefficient \cite{kendall1938}.

Sorting approaches are praised for their simplicity and low computational complexity and are thus often employed in practice. However, these approaches use heuristics that often result in suboptimal comparison choices, and in general perform worse than the methods that rely on information gain.

\paragraph{Information-gain approaches}
These methods are based on information maximization. That is, the posterior distribution of quality scores is computed and the next comparison is selected according to a utility function, e.g. Kullback-Leibler (KL) divergence \cite{kullback1951} between the current distribution and the distribution assuming any possible comparison \cite{Settles10activelearning}. This group is the most relevant to our new method. Methods listed in this section are sequential, unless stated otherwise.

A greedy Bayesian approach, \textbf{Crowd-BT}, was proposed in \cite{crowdbt2013}. The entropy for every pair of conditions is computed using the posterior distribution of each pair individually rather than jointly. The method also explicitly accounts for reliability of each annotator: scores and annotator quality are updated using an alternating optimization strategy. 

%In \cite{7938660} authors combine informativeness -- differential entropy of the probability of the correct answer in the next pairwise comparison and reliability -- probability that a certain number of comparisons offer the correct preference label for compared images in order to obtain a heuristic for active sampling. \textcolor{blue}{I would like to see a discussion after each method is explained. Why is this not suitable or why are we not considering it in the experiments? If it's not relevant to us we don't need to devote a whole paragraph to it. I remember this method was a huge mess, we need to come with a polite way of putting that. } \textcolor{red}{Let's remove it, it is way too cumbersome and we are not evaluating it.}

Authors in \cite{AAAI124747} derive the score distribution from the maximum likelihood estimation and the negative inverse of the Hessian of the log likelihood function.
Since the original implementation was not provided by the authors and our implementation suffered from numerical instability, we did not include it in our tests. 

Authors in \cite{GLICKMAN2005279,Ye2014} develop a fully Bayesian framework to compute the posterior distribution of the quality scores. \textbf{Hybrid-MST} \cite{HybridMSTRAFAL} extends this idea by selecting batches of comparisons (instead of single pairs) to maximize the information gain in the minimum spanning tree \cite{Cormen:2009:IAT:1614191} of a comparison graph. The time efficiency of the method over its predecessor is improved by computing the information gain locally \textemdash within the compared pair.

A different approach is taken by \cite{Xu2018HodgeRankWI}, where authors propose to solve a least-squares problem to elicit a latent global rating of the conditions using the Hodge decomposition of pairwise comparison data. Like other methods, the information gain is computed using the posterior of only the pair of compared conditions. We refer to this approach as \textbf{HR-active}.

%c% Mention the paper that reviewers requested.
%\paragraph{Theoretical Analysis}
%\cite{heckel2018approximate} introduces an algorithm with proof that it is optimal in number of comparisons, up to a logarithmic factor. However, the algorithm and its theoretical guarantees only cover the case of \textit{ranking}, whereas we are interested in the more general task of \textit{scaling}. Also, for many tasks the cost per comparison is very high, so even small (logarithmic) improvements are crucial.

\paragraph{Matchmaking}
A matchmaking system was proposed for gaming, together with the TrueSkill algorithm \cite{NIPS2006_3079}. The aim is to find the pairs of players with the most similar skill. The skill distribution of a pair of players is used to predict the match outcome. We refer to this approach as \textbf{TS-sampling}.

%one using the full posterior update, another using an online (approximate) posterior update for reduced computation cost; and 
\paragraph{Our Work}
In contrast to the previous work, our method (i) allows for batch and sequential modes; (ii)  estimates the posterior using the entire set of comparison outcomes that has been collected so far; and (iii) computes the utility function for a subset of pairs, saving computations without compromising on performance.

\section{Methodology}

Our algorithm consists of two main steps: (i) computing the posterior distribution of score variables \(\mathbf{r}\) using the pairwise comparisons collected; (ii) using the posterior of $\mathbf{r}$ to estimate the next comparison to be performed. In this section we first describe the score posterior estimation and then explain our active sampling algorithm. We then discuss some features to make it more computationally efficient. Pseudo-code is included in the supplementary.

\subsection{Posterior Estimation}

 \paragraph{Posterior Estimation Model} Our model is similar to Thurstone's model Case~V \cite{Thurstone1927}, with unobserved normally distributed independent random variables. However, our approach is fully Bayesian, and so instead of point value scores \(s_i\) for each condition \(o_i\), we assume that each score is a random variable \(r_i\) with distribution $r_i \sim \mathcal{N}(\mu_i, \sigma_i^2)$. Analogous to Thurstone's model, \(\mu_i\) represents the score value \(s_i\). \(\sigma_i^2\) represents the uncertainty in an estimate of \(s_i\) and is not explicitly expressed in Thurstone's model (it can be obtained, for example by bootstrapping \cite{Perez2017}). The probability that \(o_i\) is better than \(o_j\) is then given by noting that
 $r_i  - r_j \sim \mathcal{N}(\mu_i - \mu_j, \sigma^2_{ij})$, so that: 
 \begin{equation}\label{eq:distributionf} \small
 {P}(o_i \succ o_j| r_i, r_j) \triangleq \Phi \left ( \frac{\mu_i - \mu_j}{\sqrt{2} \sigma_{ij}}\right ),%\triangleq p_{ij},
 \end{equation}
where $\Phi$ is the cumulative standard normal distribution function    %$\(\mathcal{N}(0,1)\), 
and $\sigma_{ij}^2 = \sigma_i^2 + \sigma_j^2 + \beta^2$, with $\beta$ representing what is referred in the literature to as an observer/comparison noise. We further assume Thurstone Case V model in which $\beta$ is constant across all conditions. The choice of $\beta$ determines the relationship between distances in the scale and probabilities of better quality. In our experiments we set $\beta=1$. 

For a pair of compared conditions $A_t = (o_i,o_j)$ for $t \in \{1, \ldots, T\}$, where $T$ is the total number of comparisons measured so far, we denote the comparison outcome as ${y}_t \in \{-1,1\}$, where $1$ indicates that $o_i$ was preferred and $-1$ indicates that $o_j$ was preferred, with no draws allowed. 
In the inference step, we want to estimate the distribution of score variables ${\mathbf{r}}$ given $\mathbf{y}$ and $\mathbf{A} \triangleq \{A_1, \ldots, A_T\}$. The posterior distribution is: 
\begin{equation} \small
    P(\mathbf{r}|\mathbf{y}, \mathbf{A}) = \frac{P(\mathbf{y}, \mathbf{A}|\mathbf{r}) \cdot p(\mathbf{r})}{P(\mathbf{y}| \mathbf{A})},
\end{equation}
where we assume a factorizing Gaussian prior distribution over scores $p(\mathbf{r}) \triangleq \prod_{i=1}^n \mathcal{N}(r_i; {\nu}_{i}, {\alpha}^2_{i})$, $\nu_i$ and $\alpha_i^2$ being the parameters of the prior, set to $0$ and $0.5$, respectively. 
The likelihood $P(\mathbf{y}, \mathbf{A}| \mathbf{r})$ of observing comparison outcomes $\mathbf{y}$ given the ground truth scores is modelled as:
\begin{equation}\small
\label{eq:likelihood}
    P(\mathbf{y}, A|\mathbf{r}) =  \prod_{t=1}^T P(y_t, A_t | \mathbf{r}), 
\end{equation}
where 
individual likelihoods can be defined as $P(y_t,A_t|\mathbf{r}) = \mathbb{I}\left ( y_t = \sign(r_i - r_j) \right )$, i.e. equal to $1$ if the sign of $y_t$ is the same as that of the difference $r_i - r_j$ and 0 otherwise.

Although the score posterior can be written exactly via Bayes rule, the binary nature of the output factor means that the likelihood in Eq. \ref{eq:likelihood} is not conjugate to the Gaussian prior. This would lead to a non-Gaussian posterior for \(r_i\), and result in challenging, high-dimensional integrals for our information gain metric. A Gaussian approximation to messages yields a multivariate Gaussian posterior with diagonal covariance matrix, resolving both issues.

%Although the score posterior can be written exactly via Bayes rule, the binary nature of the output factor means that the likelihood in Eq. 3 is not conjugate to the Gaussian prior, and thus efficient evaluation of the \(P(o_i \succ o_j| \hat{\mathbf{r}}_{t-1})\), \(P(o_i \prec o_j| \hat{\mathbf{r}}_{t-1})\), and \(\KL\) terms will be very challenging. Applying (Gaussian approximation) expectation propagation yields a multivariate Gaussian posterior with diagonal covariance matrix, resolving the issue (and ensuring consistency with the Thurstone V model).

\paragraph{Posterior Estimation Inference}
\begin{figure}[t]
    \centering
    \includegraphics[width=0.66\linewidth]{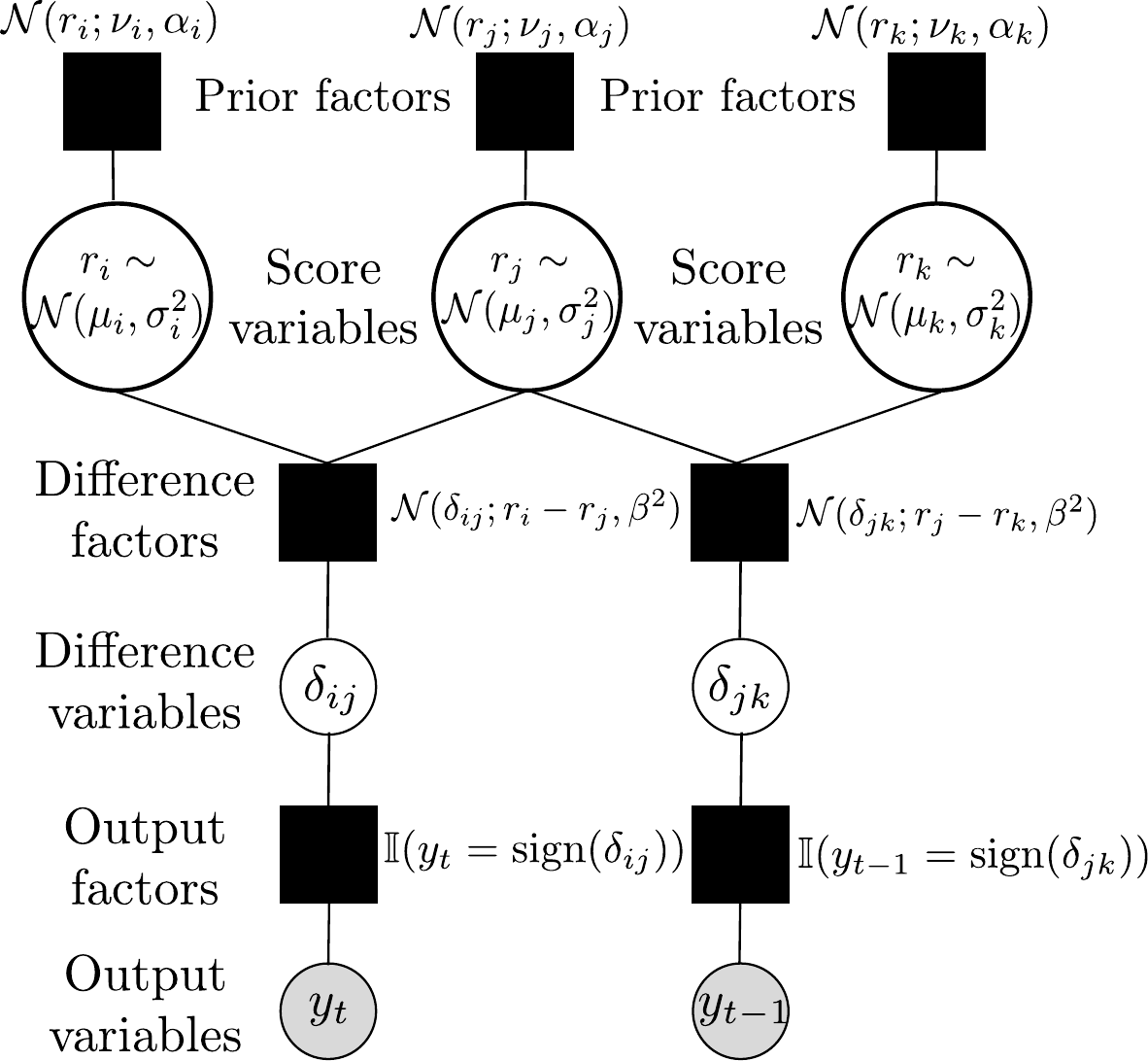}
  \caption{Factor graph for 2 comparisons of 3 conditions.}
  \label{fig:factor_graph} 
\end{figure}

Figure \ref{fig:factor_graph} shows a factor graph implementing the distribution $P(\mathbf{r}|\mathbf{y}, A)$, used as the basis for efficient inference, and inspired by TrueSkill \cite{NIPS2006_3079}. The posterior over \(r_i\) is inferred via message passing between nodes on the graph, with messages computed using the sum-product algorithm. In the general case of $n$ conditions and $T$ comparisons, we will have $n$ score variables and prior factors, $T$ difference factors, difference variables, output factors and output variables. Messages from output factors are approximated as Gaussians using expectation propagation via moment matching. %\RM{[I would remove the next sentence as it is out of context:]}The original version of TrueSkill allows for multi-person teams, draws, and time varying skills; we instead use the one person per team, no draw, no time-dependent version, as these are not necessary for our problem.

\subsection{Sampling Algorithm: ASAP} \label{IG} 

The basis of the proposed active sampling algorithm is to compute the posterior distribution over \(\mathbf{r}\) that would arise from each possible pairwise outcome in the next comparison, and then use this to choose the next comparison based on a criterion of maximum information gain. 

Several utility functions can be used to compute the expected information gain (EIG). Our choice is the commonly used Kullback-Leibler (KL) divergence \cite{kullback1951} between the prior and posterior distributions.

More specifically, our active sampling strategy picks conditions $(o_i, o_j) = A_t$ to compare in measurement $t$, such that they maximize a measure of information gain $I^{ij}_{t-1}$:
\begin{equation} \small
    A_t = \argmax_{(o_i, o_j) \in S^2, i \neq j} I_{t-1}^{ij},
\end{equation}
where $S$ is the set of all conditions and subindex $t-1$ indicates that we use all measurements collected up to the point in time $t$. For simplicity, we define $\hat{\mathbf{r}}_{t-1}$ as the estimated posterior after measurement $t-1$.

For each possible pair $A_t$, let \(P(\hat{\mathbf{r}}_{t} | y_t=+1, A_t)\) and \(P(\hat{\mathbf{r}}_{t} | y_t=-1,A_t)\) denote the updated posterior distributions (i.e. including comparison $A_t$) if $o_i$ is selected over $o_j$ ($y_t = +1$ for $A_t = (o_i, o_j)$) and vice versa. Since we cannot anticipate the outcome of the pairwise comparison, i.e. which condition will be selected, similarly to other active sampling methods \cite{HybridMSTRAFAL,Xu2018HodgeRankWI,crowdbt2013,AAAI124747,GLICKMAN2005279}, we weight the EIG with the probability of each outcome. We compute this probability using Equation \ref{eq:distributionf}, $P(o_i \succ o_j| \hat{\mathbf{r}}_{t-1})$; for condition $o_i$ selected over $o_j$ and vice versa, EIG is then defined as:
\begin{equation}\small
\begin{split}
    I^{ij}_{t-1} = & P(o_i \succ o_j| \hat{\mathbf{r}}_{t-1}) \cdot \KL \left ( P(\hat{\mathbf{r}}_t|y_t = +1, A_t)  \parallel  p(\hat{\mathbf{r}}_{t-1}) \right ) \\
    & + P(o_i \prec o_j| \hat{\mathbf{r}}_{t-1}) \cdot \KL\left ( P(\hat{\mathbf{r}}_t|y_t = -1, A_t) \parallel  p(\hat{\mathbf{r}}_{t-1}) \right ).
    \end{split}
\end{equation}

\subsection{Efficiency considerations} At every iteration $t$, the comparisons to consider is $n(n-1)/2$, where $n$ is the total number of compared conditions. The complexity of the posterior evaluation is $O(n+t)$, thus the complexity of selecting the next comparison is $O(n^2(n+t))$. This may be very costly when the number of conditions is large. Here, we discuss two modifications that reduce the computational cost, and a batch mode, which also improves the accuracy. 

\paragraph{Approximate (online) posterior estimation (ASAP-approx)}
%Since computing the full posterior after every new comparison is costly,
{In order to quantify the improvement in accuracy brought by the full posterior update, we follow the common approach, and consider the use of an online posterior update using assumed density filtering (ADF) \cite{murphy_2012}}. That is, the posterior \(\hat{\mathbf{r}}_{t-1}\) is used as the prior when computing the information gain for the \(t^{\mathrm{th}}\) comparison, allowing our algorithm to run in an online manner \cite{minka2018trueskill}. Thus, for every \(o_i\) and \(o_j\) pair, we update only the scores $r_i$ and $r_j$, resulting in $O(1)$ complexity per pair.
No additional ADF-projection step is required since expectation propagation has already yielded a Gaussian approximation to the posterior. The time complexity of selecting the next comparison is thus decreased to $O(n^2)$. However, computational efficiency comes at the cost of accuracy in posterior estimation \cite{minka2018trueskill}. We refer to the algorithm using the approximate posterior update as \textit{ASAP-approx}.

\paragraph{Selective EIG evaluations} 
Some comparisons are less informative than others \cite{Settles10activelearning}, such as conditions far apart on a scale where the outcome $y_t$ is obvious \cite{AAAI124747,GLICKMAN2005279}. Therefore we evaluate the EIG only for the most informative pairs. For that we use a simple criterion from Equation \ref{eq:distributionf} to compute the probability $Q_{ij}$ that conditions $o_i$ and $o_j$ are selected for EIG evaluation. Since Equation \ref{eq:distributionf} is the probability that condition $o_i$ is better than $o_j$, to identify obvious outcomes we set $Q_{ij} = min(p_{ij},p_{ji})$. Thus, the probability is large when the difference between the scores and their standard errors are small. To ensure that at least one pair including $o_i$ is selected, we scale  $Q_{ij}$ per condition, i.e. $Q^*_{ij} = \frac{Q_{ij}}{\max_{\forall j}(Q_{ij})}$.

%We use the matchmaking equation from \cite{NIPS2006_3079} to compute the likelihood that conditions $o_i$ and $o_j$ are selected for EIG evaluation:
%\begin{equation} \small
%q_{ij}(A_t,\mathbf{r}) =\sqrt{\frac{2\beta^2}{2\beta^2+\sigma_i^2+\sigma_j^2}}\cdot \exp\left({-\frac{(\mu_i-\mu_j)^2}{2(2\beta^2+\sigma_i^2+\sigma_j^2)}}\right).
%\end{equation}
%The first term of the multiplication is large when standard errors of the scores are small. The second term is large when both the difference between the scores and their standard errors are small. This matchmaking equation is derived from the definition of the probability of scores being the same up to a draw margin $\epsilon$ (i.e. scores being close in the scale, $\mu_i-mu_j \leq \epsilon$) in the limit $\epsilon \rightarrow 0$, therefore adapting to different scales. To ensure that at least one pair including $o_i$ is selected, we scale $q$ per condition, i.e. $q^{*}_{ij} = \frac{q_{ij}}{\max_{\forall j}(q_{ij})}$.

\paragraph{Minimum spanning tree for the batch mode} When a sampling algorithm is in the sequential mode, one pair of conditions is scheduled in every iteration of the algorithm. However, selecting a batch of comparisons in a single iteration of the algorithm is computationally more efficient and can yield superior accuracy \cite{HybridMSTRAFAL}. To extend our algorithm to the batch mode, we treat pairwise comparisons as an undirected graph. Vertices are conditions, and edges are pairwise comparisons. We follow the approach from \cite{HybridMSTRAFAL} where the minimum spanning tree (MST) is constructed from the graph of comparisons. The MST is a subset of the edges connecting all the vertices together, such that the total edge weight is minimal. The edges of our graph are weighed by the inverse of the EIG, i.e. for an edge $E_{ij}$ connecting conditions $A_i$ and $A_j$ the weight is given by $w(E_{ij}) = \frac{1}{I^{ij}}$. $n-1$ pairs are selected for the MST, allowing us to compute the EIG every $n-1$ iterations, greatly improving speed. Since each condition is compared at least once within our batch, detrimental imbalanced designs \cite{JMLR:v18:16-206}, where a subset of conditions is compared significantly more often than the rest, are eliminated.

\section{Evaluation}\label{sec:experiments}
To assess different sampling strategies, we run Monte Carlo simulation on synthetic and real datasets. Spearman rank ordering correlation coefficient (SROCC) and root-mean-squared Error (RMSE) between the ground truth and estimated scores are used for performance evaluation. We report our results as multiples of standard trials, where \textbf{1 standard trial} corresponds to $n(n-1)/2$ measurements (the number of possible pairs for $n$ conditions). For clarity, we present RMSE on a log-scale, and SROCC after a Fisher transformation ($y' = \arctanh(y)$). The same method, based on approximate message passing, was used to produce the scale from pairwise comparisons for each method. We verified that the scaled results are consistent with the MLE-based method from \cite{Perez2017}.

\subsection{Algorithms compared}
We implement and compare different active sampling strategies using original authors' codes where possible: AKG \cite{chen2016}, Crowd-BT \cite{crowdbt2013}, HR-active \cite{Xu2018HodgeRankWI} and Hybrid-MST \cite{HybridMSTRAFAL}. Our own implementation was used for Quicksort \cite{Maystre2017}, Swiss System \cite{Ponomarenko2015}, and TS-sampling \cite{NIPS2006_3079}. 

\subsection{Simulated Data}
In order to control the ground truth distribution underlying the data, we first run a Monte Carlo simulation with synthetic data. In the simulation, we use \(P(o_i \succ o_j | r_i, r_j)\) from Equation \ref{eq:distributionf} to draw $y_t$ for comparison at trial $t$ between conditions \(o_i\) and \(o_j\), which are determined by each algorithm. 
We note that the strongest influence on the results is the proximity of compared conditions in the target scale. When conditions have comparable scores, they are confused more often in comparisons, whereas when conditions are far apart in the scale they are easily distinguished, resulting in different performances for sampling methods. Hence, we consider 3 scenarios for 20 conditions with scores $\mathbf{s}$ sampled uniformly from: (i) large range $[0,20]$ (scores far apart); (ii) medium range $[0,5]$; (iii) small range $[0,1]$ (scores close together). 
Results for larger numbers of conditions are given in Section~\ref{sec:large-scale}. We run the simulation 100 times for comparisons ranging from 1 to 15 standard trials.

\begin{figure}[t]
    \centering  

  \subfloat[Saved evaluations]{%
       \includegraphics[width=0.47\linewidth]{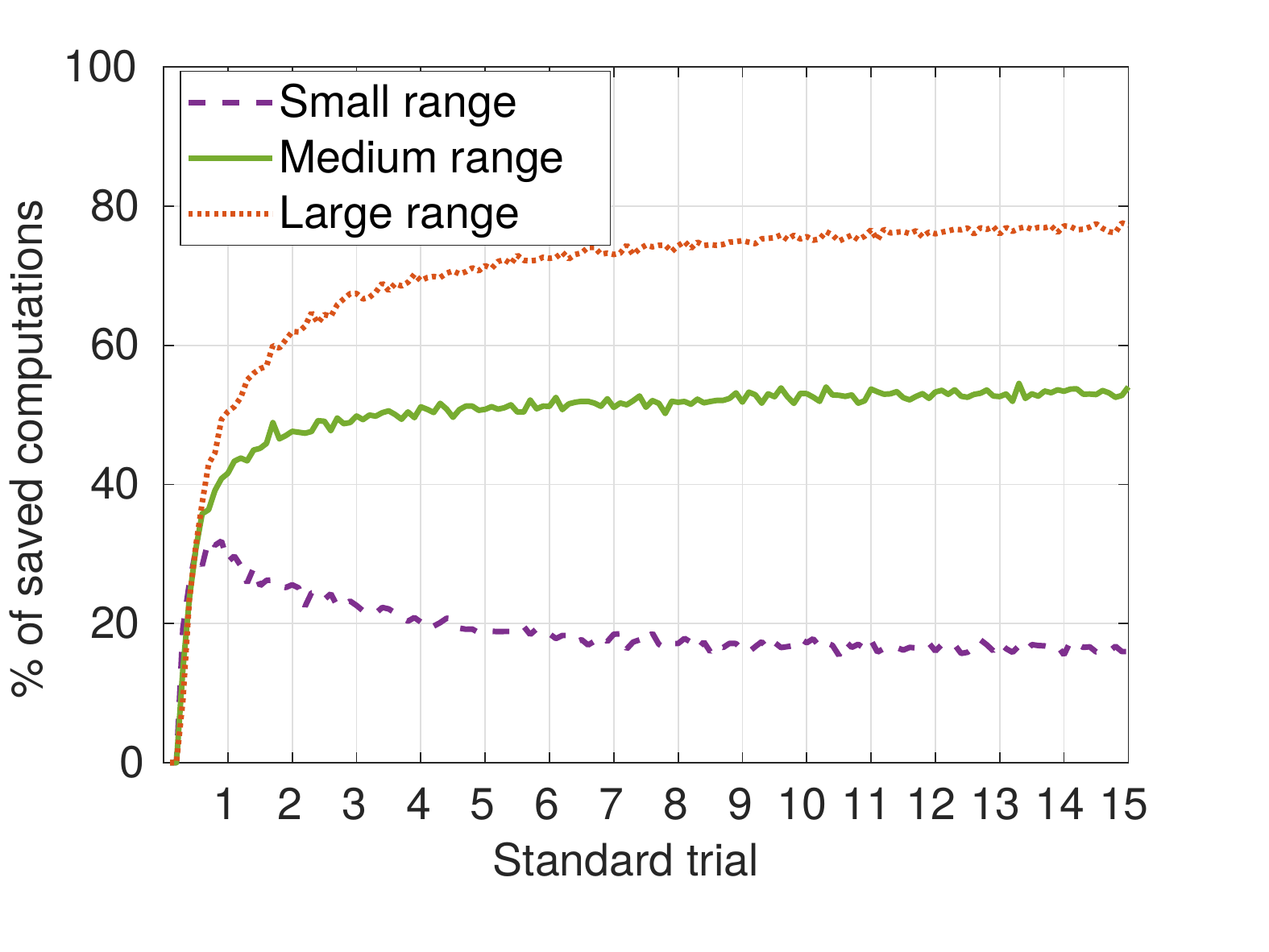}\label{fig:selective_computations_saved_iterations_left}
       }
  \subfloat[Probability of EIG evaluation]{%
        \includegraphics[width=0.47\linewidth]{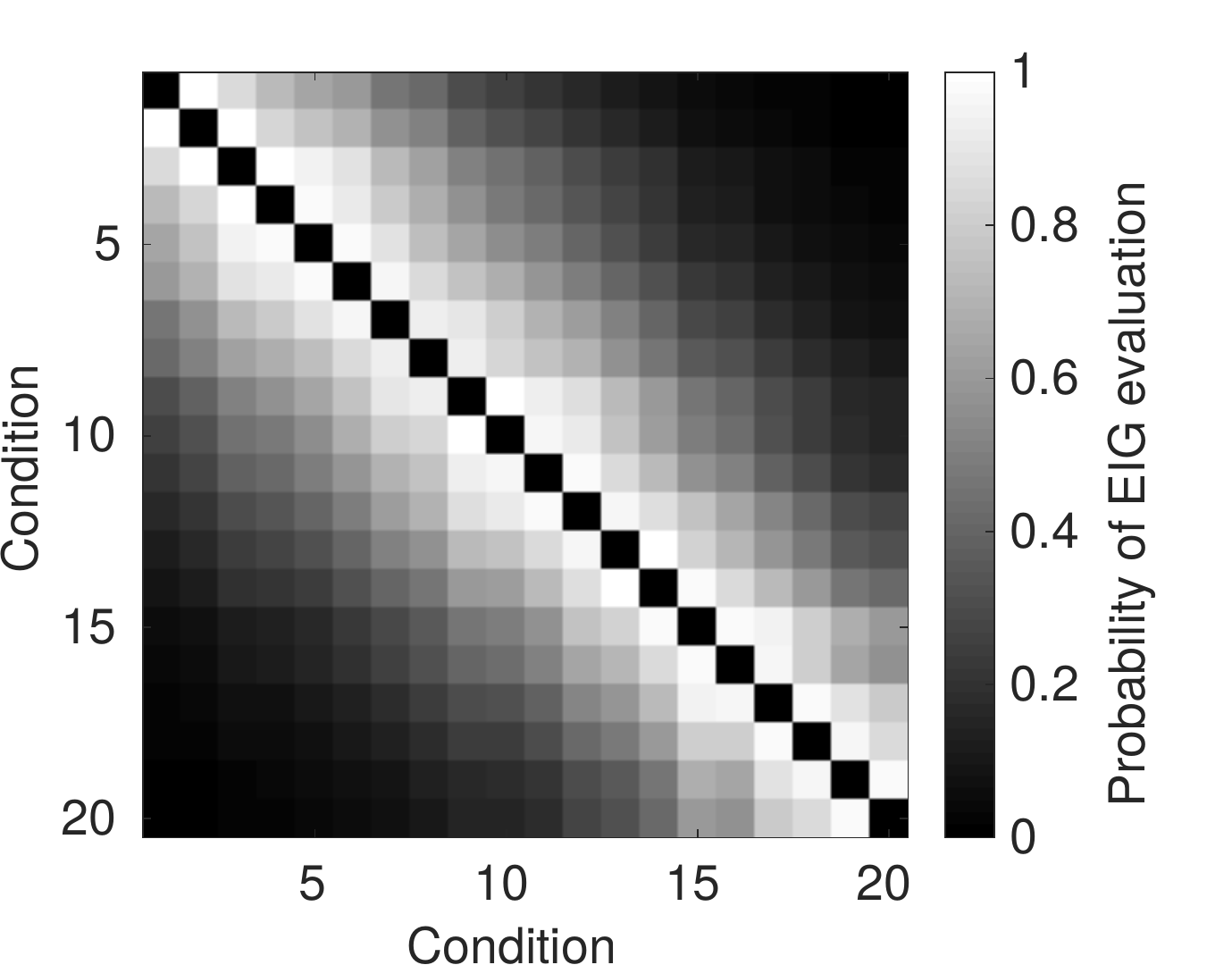}\label{fig:selective_computations_saved_iterations_right}
        }\\
            \subfloat[Performance with and without selective EIG]{%
       \includegraphics[width=0.98\linewidth]{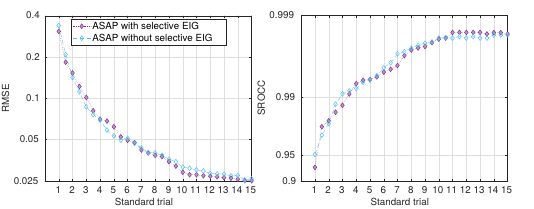}
       \label{fig:eig_perf}}
  \caption{(a) Percentage of saved evaluations with selective EIG evaluations; (b) probability of EIG evaluation after 10 standard trials for medium range; and (c) RMSE and SROCC with and without selective EIG; }
  \label{fig:selective_computations_saved_iterations} 
\end{figure}

\paragraph{Selective EIG evaluations} Figure \ref{fig:selective_computations_saved_iterations_left} shows the proportion of saved evaluations with selective EIG computations. Since we initialize our algorithm with all scores set to 0, all possible pairs have their EIG computed at first (0 standard trials in the plot), as all conditions are close to each other. As more data are collected, conditions move away from each other on the scale and the EIG is computed for a subset of pairs only. Computational saving is greater for large-range simulations than for small-range simulations. In small-range simulations, conditions first move away from each other, as in the first few iterations their relative distances are likely to be overestimated, decreasing the overall number of computations; however, with more measurements the conditions move closer, and the proportion of saved evaluations decreases. Figure \ref{fig:selective_computations_saved_iterations_right} shows the probability of the EIG being evaluated after 10 standard trials for 20 conditions sampled from the medium range. For visualization purposes, conditions were ordered ascending in the quality scale. Pairs of conditions along the diagonal, i.e.\ close in the scale, have a higher chance of their EIG being computed. Figure \ref{fig:eig_perf} shows performance of ASAP with and without selective EIG evaluations. Thus, selectively evaluating EIG greatly reduces the number of computations while maintaining the same accuracy measured in RMSE and SROCC. In the following sections, we only present the results with selective EIG computations.

\begin{figure}[t]
   \centering
  \includegraphics[width=0.97\linewidth]{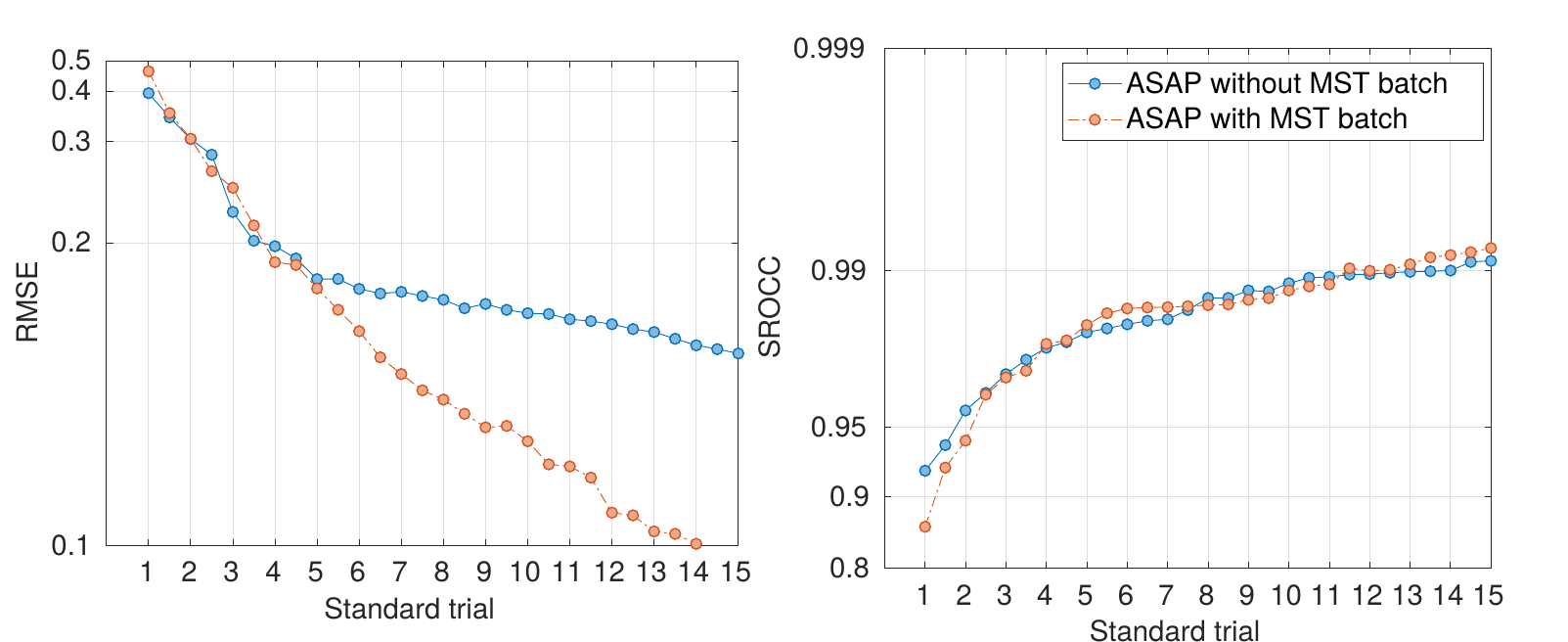}
  \caption{Simulation with 20 conditions sampled from the medium range with and without MST. We observe similar pattern for conditions sampled from small and large ranges.}
  \label{fig:MST_batch} 
\end{figure}

\paragraph{Minimum spanning tree for the batch mode} Figure \ref{fig:MST_batch} shows the results of ASAP with and without batch mode for medium-range simulations. Without MST batch mode, the method is likely to result in an imbalanced sampling pattern, where certain conditions are compared significantly more often than others. This has a detrimental effect on performance, deteriorating the results with growing number of comparisons \cite{JMLR:v18:16-206}. Below, we only present results with MST batch mode.

\begin{figure}[t]
    \centering
    \includegraphics[width=\linewidth]{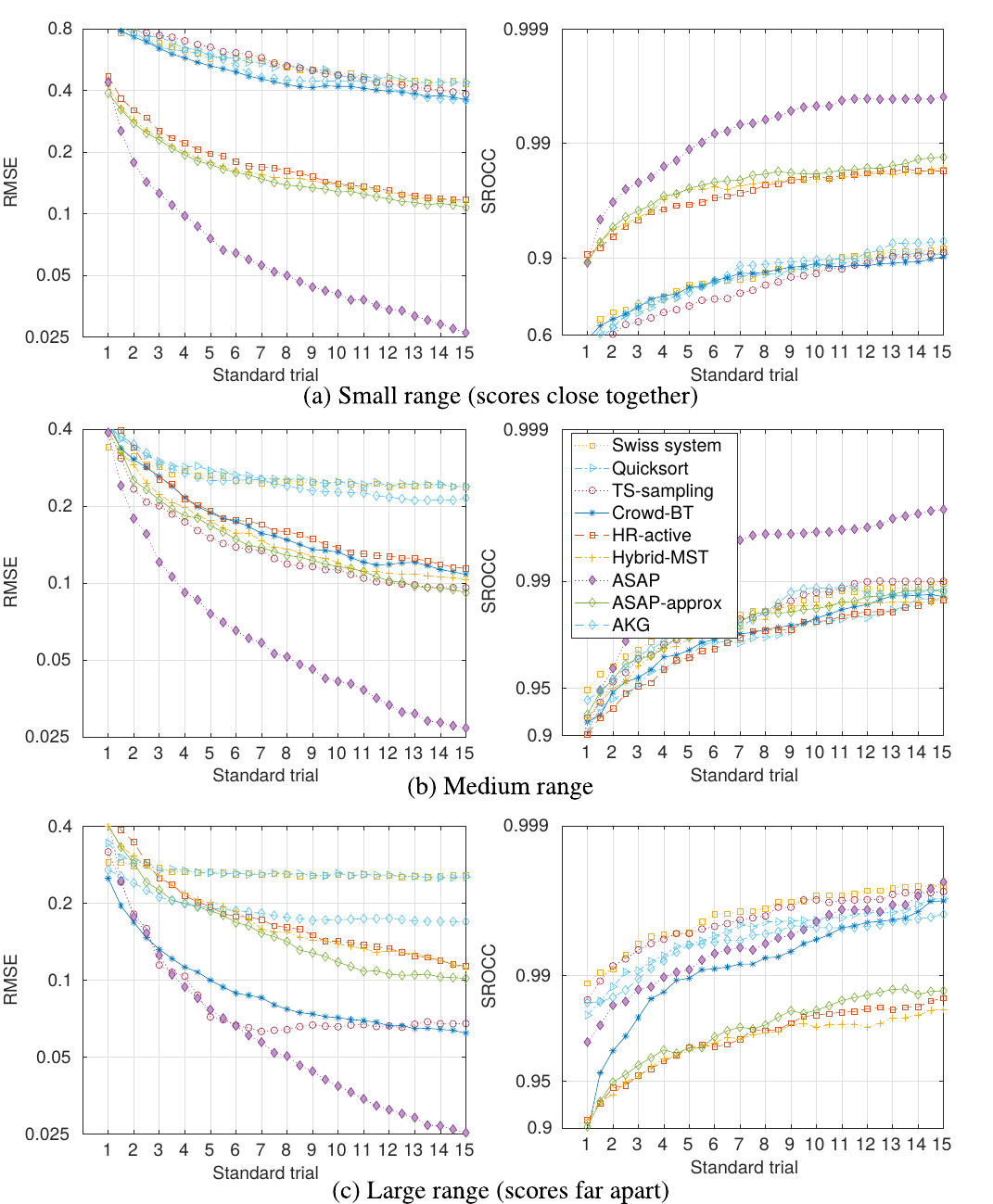}
  \caption{Simulation results with 20 conditions for the compared sampling strategies.}
  \label{fig:samplingsimulation} 
\end{figure}

\paragraph {Simulation results} Figure \ref{fig:samplingsimulation} shows the results of the simulation for the implemented strategies. 
At all tested ranges, EIG-based methods have lower RMSE, and therefore higher accuracy, than the sorting methods (Quicksort and the Swiss System).  While TS-sampling and Crowd-BT have good accuracy for the large range, these are among the worst methods for the small range. ASAP-approx exerts performance similar to the methods with online posterior update, however offers a modest but consistent improvement in accuracy over Hybrid-MST and HR-active. Of all tested methods, ASAP, employing full posterior update, is the most accurate by a substantial margin and across all ranges.

For SROCC, EIG-based methods do not show a clear advantage over sorting methods; however, it should be noted that EIG-based methods are designed to optimize for RMSE rather than ranking. Even so, ASAP still performs the best for small and medium range simulations, and one of the best for large range, reaching SROCC of 0.99 within five standard trials. It should be noted, however, that the problem of ordering conditions from the large range is trivial and the best methods compete at 0.99+ SROCC levels (almost perfect ordering). Because of the poor performance of the sorting-based methods, we do not consider them in the following experiments.

\subsection{Real Data}
We validate the performance of sampling strategies on two real-world datasets: i) Image Quality Assessment (IQA) LIVE dataset \cite{Sheikh2006b}, with pairwise comparisons collected by \cite{Ye2014}; and ii) Video Quality Assessment (VQA) dataset \cite{Xu2011}. Each dataset contains complete and balanced matrices of pairwise comparisons, with each condition compared to every other condition the same number of times. The empirical probability of one condition being better than another is obtained from the measured data and used throughout the simulation. We compute RMSE and SROCC between scores produced by each method, and scores obtained by scaling the original matrices of all comparisons.

\paragraph{IQA dataset} 
To allow multiple runs of the Monte Carlo simulation, we randomly select 40 conditions from the 100 available. In the original matrix, each condition is compared 5 times with each other (5 standard trials), yielding 24750 comparisons. 

\begin{figure}[t]
    \centering
\includegraphics[width=\linewidth]{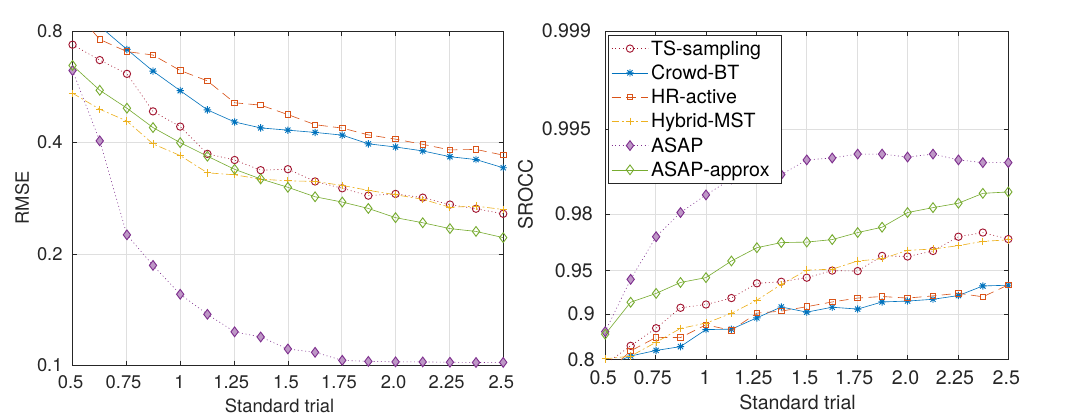}
  \caption{Compared sampling strategies on LIVE dataset.}
  \label{fig:live_sampling} 
\end{figure}

Figure \ref{fig:live_sampling} shows the results. The performance trends are consistent with the results for the simulated data for the medium range. ASAP has the best performance both in terms of SROCC and RMSE. It is followed by ASAP-approx, Hybrid-MST, and TS-sampling, each having roughly the same performance in terms of both RMSE and SROCC. Crowd-BT and HR-active have the worst performance in terms of both RMSE and SROCC. 

\paragraph{VQA dataset}
The dataset contains 10 reference videos with 16 distortions. Each $16\times16$ matrix contains 3840 pairwise comparisons, i.e.\ each pair was compared 32 times.
\begin{figure}[t]
    \centering
       \includegraphics[width=\linewidth]{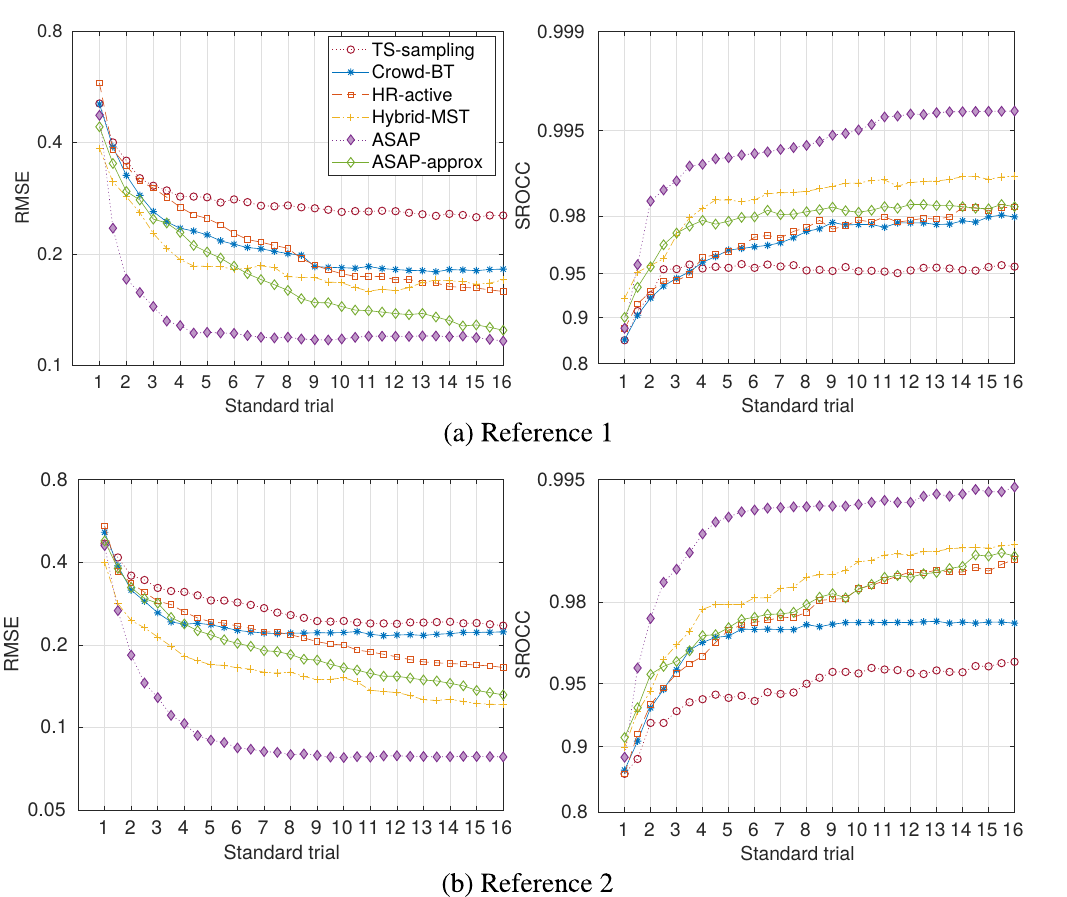}
  \caption{Compared sampling strategies on VQA dataset. }
  \label{fig:vqa} 
\end{figure}

Figure \ref{fig:vqa} shows the results of running simulations on the first two reference videos. The performance trends are again, in general, consistent with the results for the simulated data sampled from the medium range, except that TS-sampling performs substantially worse, and Hybrid-MST outperforms ASAP-approx for small numbers of trials. ASAP consistently outperforms other methods. The results for the remaining eight reference videos are given in the supplementary. % follow the same trend and These are omitted from the paper due to space restrictions.

%ASAP has the best performance in terms of SROCC and RMSE across all reference videos. It is followed by Hybrid-MST and ASAP-approx. TS-sampling has the worst performance with HR-active and Crowd-BT having similar slightly better performance performance. 

\subsection{Large Scale Experiments}
\label{sec:large-scale}
\begin{figure}[t]
    \centering
    \includegraphics[width=0.97\linewidth]{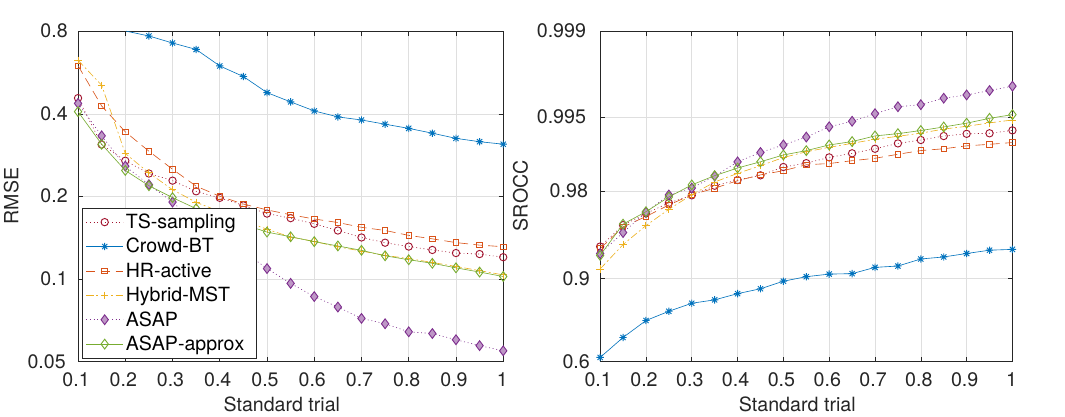}
  \caption{Large scale experiment simulation with 200 conditions sampled from the medium range.}
  \label{fig:large_scale_experiment} 
\end{figure}

It is often considered that 15 standard trials is the minimum requirement for FPC to generate reliable results \cite{itu910,itu500}, however, this is rarely feasible in practice. Real-world large-scale datasets barely reach 1 standard trial. To make experiments with large number of conditions feasible, individual reference scenes or videos are often measured and scaled independently, missing important cross-content comparisons. However, the lack of cross-content comparisons yields less accurate scores \cite{zerman:hal-01654133}. Active sampling techniques, such as ASAP, should accurately measure a large number of conditions, while saving substantial amount of experimental effort. To test such a scenario, we simulate the comparison of 200 conditions with scores distributed in the medium range. The results, shown in Figure \ref{fig:large_scale_experiment}, demonstrate that even with a small number of standard trials ASAP outperforms existing methods; it is followed by ASAP-approx and Hybrid-MST. 

%In image and video quality experiments, quality scores are often measured and scaled independently. However, such an experiment design misses [lacks?] important cross-content comparisons, resulting in less accurate measurements of quality scores. A FPC would be ideal, but it is not feasible to collect the minimum 15 standard trials of data required [18,2]; real-world data sets barely reach 1 standard trial. 

% rafal: I would argue that these standard trials are not correctly computed
%, e.g. image quality assessment dataset TID2013, containing 3,000 conditions \cite{Ponomarenko2015} contains 0.116 standard trials. Another large scale image quality dataset, PieApp  \cite{Prashnani_2018_CVPR} containing 10,000 conditions has 0.76 standard trials. To collect 1 standard trial for a dataset with 3,000 conditions, one requires 4,498,500 comparisons, which, assuming an unrealistically modest 1 second decision time for a human participant, converts into $~1250$ hours of experiments. For that reason we stress the importance of the method performance for the large number of conditions and low trial number. Figure \ref{fig:large_scale_experiment} shows the performance of sampling strategies on 200 conditions sampled from the medium range. Our method shows performance similar to Hybrid-MST, however slightly outperforms it for smaller standard trial numbers. HR-active follows in performance.}

\subsection{Running Time and Experimental Effort}
\begin{figure}[t]
  \centering
  \includegraphics[width=0.95\linewidth]{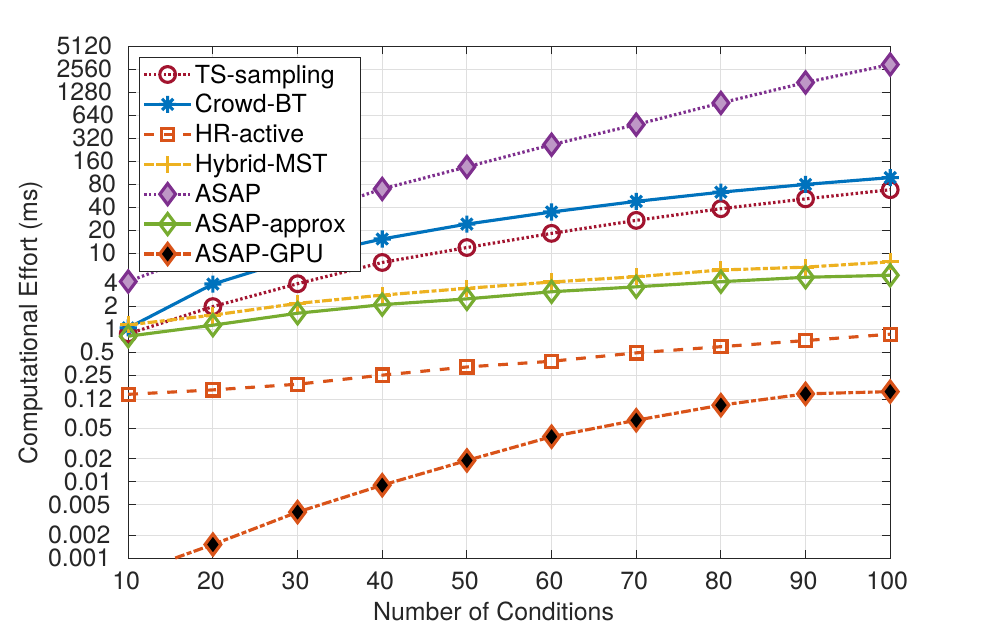}
  \caption{Average time to select the next comparison for a varied number of conditions and 5 standard trials.}
  \label{fig:time_trend}
\end{figure}

A practical active sampling method must generate new samples in an acceptable amount time. Hence, in Figure \ref{fig:time_trend} we plot the time taken by each method as the number of conditions grows. The reported times are for generating a single pair of conditions, assuming that 5 standard trials have been collected so far. CPU times were measured for MATLAB R2019a code running on a 2.6GHz Intel Core i5 CPU and 8GB 1600MHz DDR3 RAM. GPU time was measured for Pytorch 1.4 with CUDA 9.2, running on GeForce GTX1080. %\RM{Is there a reason why you collected the timings on a laptop? Comparing 1080 to a laptop's CPU looks strange?}. 
We omit sorting methods as they do not offer sufficient accuracy. Although ASAP is the slowest method when running on a CPU, it can be effectively parallelized on a GPU and deliver the results in a shorter time than other methods running on a CPU. %Therefore, the computation time is not a limitation of ASAP.

%\textcolor{red}{We also include a GPU implementation of ASAP in Python, which we ran on a GTX 1080 GPU. Unsurprisingly, ASAP-GPU is the fastest method. HR-active and ASAP-approx follow. ASAP is the slowest. However, even for 100 conditions and 5 standard trials, the computation time of about 3 seconds could be acceptable.}

\begin{figure}[t]
  \centering
  \includegraphics[width=0.95\linewidth]{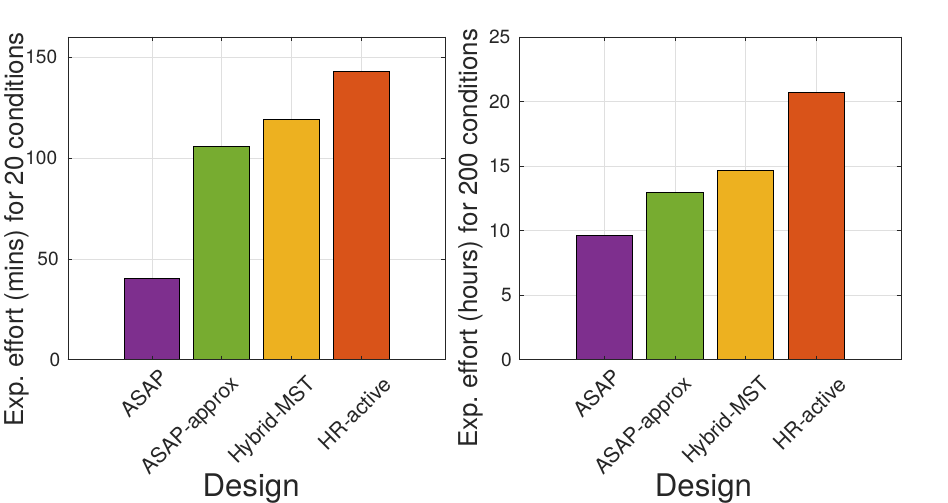}
  \caption{Experimental effort (amount of time, assuming 5 second decision time, required to reach 0.15 RMSE) for experiments with 20 and 200 conditions.% and time (average time to generate the next pair of conditions to compare) for simulations with 20 (left) and 200 (right) conditions. 
  }
  \label{fig:time_plots}
\end{figure}

In Figure~\ref{fig:time_plots} we show the experimental effort required to reach an acceptable level of accuracy for 20 and 200 conditions, where we define experimental effort as the time required to reach an RMSE of 0.15. We assume that each comparison takes 5 seconds, which is typical for image quality assessment experiments \cite{2019TIP, Ponomarenko2015}.  %To illustrate where each method stands in terms of accuracy in practical applications, we find the experimental effort (time for 20 and 200 conditions respectively) to reach the RMSE of 0.15 or less. 
%The y-axis shows the experimental effort averaged across the three simulated score ranges for 20 conditions (methods not reaching the 0.15 threshold are not listed). The x-axis shows the experimental effort for 200 conditions.
ASAP offers the biggest saving in experimental effort for both small and large scale experiments. In an experiment with 200 conditions ASAP achieves an accuracy of 0.15 RMSE in 0.355 standard trials. The total experimental time is thus 9.8h (7065 comparisons), which is significantly better than the 14.6h (10550 comparisons) for Hybrid-MST. Similarly, for 20 conditions the entire experiment would take 40 min for ASAP and 120 min for Hybrid-MST to reach the same accuracy of score estimates. 
For experiments with longer comparison times (e.g.\ video comparison) or high comparison cost (e.g.\ medical images) ASAP's advantage is even greater. %Although the computation cost is higher for ASAP compared to other methods, selecting a batch of 200 comparisons is computationally feasible for real life experiments - under 3 mins (under 15 mins for a batch of 500 comparisons). %Furthermore, for large scale experiments ASAP can be easily parallelizable.

%\RM{I would de-emphasize ASAP-approx and remove this paragraph. Why would anyone want spend more time running experiments if a better method exist? } ASAP-approx trades accuracy and experimental effort for lower compute time. While the improvement of ASAP-approx over the state-of-the-art is marginal in terms of accuracy, its computational effort improvement is substantial. ASAP-approx requires 20-30\% less time to schedule a comparison than for comparable accuracy in Hybrid-MST. %The other methods tend to be less accurate than the two proposed, with more accurate methods requiring more computation time.

%20 conditions: HR-active 84 min, Hybrid-MST  71 min, ASAP 23 min  ASAP-approx 62 min 
%200 conditions: TS-sampling 10.5h, HR-active 11.9h Hybrid-MST 7.9h, ASAP 9.2h, ASAP-approx 7.4h

%In this section we analyse time performance and compare the combination of both experimental effort and computation time required to select the most informative next comparison. The experimental effort is measured as the number of standard trials needed for the algorithm to achieve 0.15 RMSE for the conditions sampled from the medium range. All experiments were conducted with MATLAB R2019a on a 2GHz Intel Core i5 CPU and 8GB 1600MHz DDR3 RAM.

%Figure \ref{fig:time_plots} shows the result of this experiment. Figure \ref{fig:time_trend} shows the time trend for the compared methods. Crowd-BT has poor experimental performance and  large computational cost. HR-active is the fastest method, however is only slightly better than Crowd-BT on some occasions. ASAP is the slowest method having a drawback of the computational time growing with the number of comparisons, however has the best experimental performance. ASAP-approx on the other hand has the better than average performance, outperforming Hybrid-MST on simulated data and lacking slightly behind on VQA datasets. Both Hybrid-MST and ASAP-approx have a moderate computational cost. TS-sampling has the worst performance on the the real world data, and average performance on the simulated.
%Sorting based sampling strategies have the worst experimental performance, however they are fast.

\section{Conclusions} \label{sec:conc}

In this paper, we showed the importance of choosing the right sampling method when collecting pairwise comparison data% for subjective quality assessment
, and proposed a fully Bayesian active sampling strategy for pairwise comparisons -- ASAP. 

Commonly used sorting methods perform poorly compared to the state-of-the-art methods based on the EIG, and even EIG-based methods are sub-optimal, as they rely on a partial update of the posterior distribution. ASAP computes the full posterior distribution, which is crucial to achieving accurate EIG estimates, and thus the accuracy of active sampling. Fast computation of the posterior, important for real-time applications, was made possible by using fast and accurate factor graph approach, which is new to the active sampling community. In addition, ASAP only computes the EIG for the most informative pairs, reducing the computational cost of ASAP by up to 80\%, and selects batches using a minimum spanning tree method, allowing to avoid imbalanced designs.

We recommend ASAP, as it offered the highest accuracy of inferred scores compared to existing methods in experiments with real and synthetic data. The computational cost of our technique is higher than for other methods in the CPU implementation, but is still in the range that makes the technique practical, with a substantial saving of experimental effort. For large-scale experiments, in GPU implementation ASAP offers both accuracy and speed. %For extremely compute-intensive experiments, or where GPU is not available, we recommend ASAP-approx, which uses assumed density filtering for online posterior updates. ASAP-approx achieves the accuracy of the state-of-the-art methods with only 70\% computational cost.

%For very large set of conditions, we suggest using Hybrid-MST, or an approximate version of our technique, ASAP-approx.

%% Old version
%In this work we presented two active sampling algorithms for pairwise comparisons based on expected information gain maximisation: ASAP and its  computationally efficient modification -- ASAP-approx. Our algorithm uses an expectation propagation to compute the full posterior of the score distribution and provides the most accurate solution both in terms of ordering and scaling the conditions. ASAP-approx encapsulates the trade off between the computational efficiency and performance, producing better than average results, however at a moderate computational cost. Both strategies have a batch mode extension, allowing for better experimental efficiency and more accurate results. We also improve the computational efficiency by computing the information gain criterion only for the pairs of conditions.

\section*{Acknowledgments}
This project has received funding from EPSRC research grant EP/P007902/1, from the European Research Council (ERC) under the European Union’s Horizon 2020 research and innovation programme (grant agreement N$^\circ$ 725253 (EyeCode), and from the Marie Sk{\l}odowska-Curie grant agreement N$^\circ$ 765911 (RealVision).


%\section*{Acknowledgment}




\bibliographystyle{IEEEtran}
\bibliography{IEEEabrv,ref}
\pagebreak

%\begin{thebibliography}{1}

%H.~Kopka and P.~W. Daly, \emph{A Guide %to \LaTeX}, 3rd~ed.\hskip 1em plus

%\end{thebibliography}

\end{document}

\title{Supplementary material for: ``Active Sampling for Pairwise Comparisons via Approximate Message Passing and Information Gain Maximization"}

\author{\IEEEauthorblockN{Aliaksei Mikhailiuk\IEEEauthorrefmark{1}, Clifford Wilmot\IEEEauthorrefmark{1}, Maria Perez-Ortiz\IEEEauthorrefmark{3}\IEEEauthorrefmark{1}, Dingcheng Yue\IEEEauthorrefmark{1}, Rafa{\l} K. Mantiuk\IEEEauthorrefmark{1}}
\IEEEauthorblockA{\IEEEauthorrefmark{1}Department of Computer Science University of Cambridge
Cambridge, United Kingdom
}
%\IEEEauthorblockA{ \IEEEauthorrefmark{2}Unaffiliated, 
%London, United Kingdom,
%email: cliffw29@outlook.com
%}
\IEEEauthorblockA{\IEEEauthorrefmark{3}Department of Computer Science, University College London, 
London, United Kingdom\\
Email: am2442@cam.ac.uk, cliffw29@outlook.com, maria.perez@ucl.ac.uk, \{dy276, rafal.mantiuk\}@cam.ac.uk}
}

%Telephone: (800) 555--1212, Fax: (888) 555--1212}


\maketitle

\section{Introduction}
We include pseudo-code for the ASAP algorithm, sampling patterns of the ASAP, the distribution of the RMSE and SROCC for the conditions sampled from the large range (bottom row of Figure 4 in the main paper) and the results for the 8 reference videos of the Video Quality Assessment (VQA) dataset.

\subsection{Pseudo-code}
We provide a detailed description of the ASAP method in Algorithm \ref{alg:algorithm}. The notation in the algorithm corresponds to that used in the main paper. The algorithm requires as input a list of comparisons performed so far, $\mathbf{y}$ and a matrix with the probabilities of the EIG being computed \textemdash required for the selective evaluations \textemdash $\mathbf{Q}$. The algorithm outputs a batch of pairs to be compared $\mathbf{C}$ and an updated probability of being selected for the EIG evaluations, $\hat{\mathbf{Q}}$.

\begin{algorithm*}[t]

\SetAlgoLined
\textbf{Input}: $\mathbf{y}, \mathbf{Q}$ \\
\textbf{Output}: $\mathbf{C}, \hat{\mathbf{Q}}$ \\

\emph{\# Calculate the posterior for the given state of the comparison matrix}\\
$\boldsymbol{\mu}, \boldsymbol{\Sigma}$ = approxPosterior($\mathbf{y}$)

\emph{\# Iterate over the rows of the expected information gain matrix $\mathbf{I}$}\\
\For{$i\leftarrow 1$ \KwTo $n$}{

\emph{\# Iterate over the columns of the expected information gain matrix $\mathbf{I}$}\\
\For{$j\leftarrow 1$ \KwTo $(i-1)$}{

\emph{\# Probability of selecting $o_i$ over $o_j$}\\
$p_{ij}={P}(o_i \succ o_j| r_i, r_j) = \Phi \left ( \frac{\mu_i - \mu_j}{\sqrt{2} \sigma_{ij}}\right )$

\emph{\# Selective EIG evaluations with roulette}\\
\If{$Q_{ij}>U[0,1]$}{

  \emph{\# Posterior given all comparisons and assuming $o_i$ is selected over $o_j$ }\\
  $\mu_{ij}, \Sigma_{ij}$ = approxPosterior($\mathbf{y}$; $o_i\succ o_j$)
  
  \emph{\# Posterior given all comparisons and assuming $o_j$ is selected over $o_i$ }\\
  $\mu_{ji}, \Sigma_{ji}$ = approxPosterior($\mathbf{y}$; $o_j\succ o_i$)
  
  \emph{\# KL divergence between current distribution and distribution assuming the two possible outcomes}\\
  $KL_{ij}$ = KLDivergence($\mathcal{N}(\boldsymbol{\mu}, \boldsymbol{\Sigma}), \mathcal{N}(\mu_{ij}, \Sigma_{ij})$)

  $KL_{ji}$ = KLDivergence($\mathcal{N}(\boldsymbol{\mu}, \boldsymbol{\Sigma}), \mathcal{N}(\mu_{ji}, \Sigma_{ji})$)
  
  \emph{\# Weighted information gain}\\
  $I_{ij} = p_{ij}\times KL_{ij} + (1-p_{ij})\times KL_{ji}$
  
  }
  
  \emph{\# Update the probability of being selected for the comparison}\\
  $\hat{Q}_{ij} =min({p}_{ij} ,1-{p}_{ij}) %=\sqrt{\frac{2\beta^2}{2\beta^2+\sigma_i^2+\sigma_j^2}}\cdot \exp\left({-\frac{(\mu_i-\mu_j)^2}{2(2\beta^2+\sigma_i^2+\sigma_j^2)}}\right)
  $
  }
  
  \emph{\# Scale $q$ per condition}\\
  $\hat{Q}_{ij} = \frac{\hat{Q}_{ij}}{\max_{\forall j}(\hat{Q}_{ij})},~\forall j$
  
  %$\hat{Q}_{ij} = min(\hat{Q}_{ij} ,1-\hat{Q}_{ij})$
  
  }
  
  \emph{\# Make the EIG matrix symmetric and find reciprocal of each entry}\\
  $I = \mathbf{1}/(I+I^T)$
 
  \emph{\# Create the minimum spanning tree from the matrix $\mathbf{I}$}\\
  G = minspantree($\mathbf{I}$)
        
 \emph{\# Nodes connected by an edge are pairs of conditions to compare}\\
 $\mathbf{C}$ =  getConnectedNodes(G)
 
 \emph{\# Note if batch mode is not used pairs to compare are selected by  $\mathbf{C}$ = argmax($I_{ij}$)}
 
 \caption{ASAP}
 \label{alg:algorithm}
\end{algorithm*}

%\subsection{Selective EIG evaluations}
%Selective EIG evaluations do not influence the accuracy of ASAP, as seen from Figure \ref{fig:eig}. Pairs selected for the EIG evaluation are likely to be selected by the sampling algorithm if EIG is not used, thus selective EIG evaluations are preserving the informative pairs and discarding less informative ones.

%\begin{figure}[ht!]
%    \centering
%\includegraphics[width=\linewidth]{./figures/EIG.pdf}
%  \caption{Performance of ASAP with and without selective EIG evaluations for conditions sampled from the medium range. We observe similar behaviour for conditions sampled from small and large ranges.}
 % \label{fig:eig} 
%\end{figure}
\subsection{Sampling patterns}
\begin{figure}[h]
    \centering
\includegraphics[width=\linewidth]{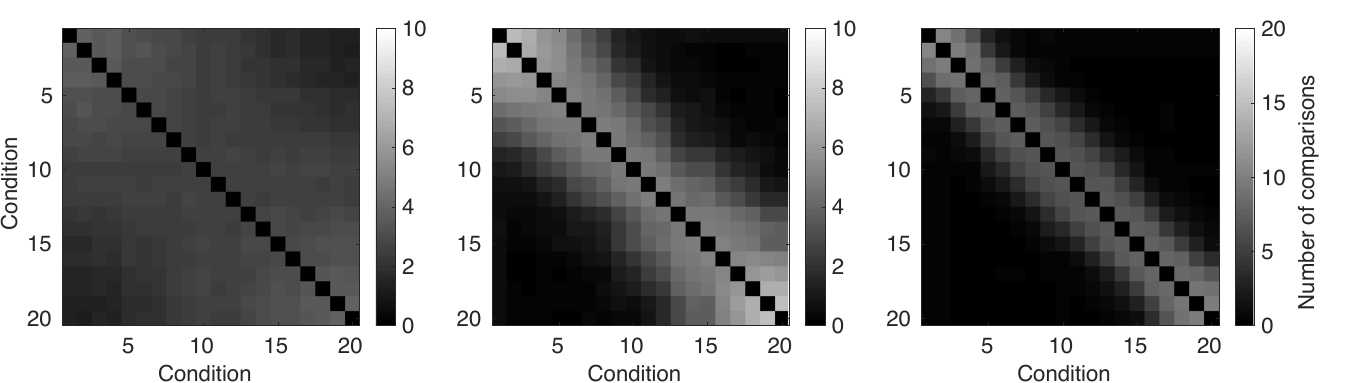}
  \caption{Heatmaps for 3 standard trials of comparisons of 20 ordered conditions from small, medium and large ranges}
  \label{fig:heatmaps} 
\end{figure}

 In order to better understand which conditions are favored by ASAP, we produce heatmaps of pairings for conditions sampled from the small, medium and large ranges. We use 3 standard trials. The heatmaps are given in Figure \ref{fig:heatmaps}. For better visualization, conditions are ordered ascending in their ground truth scores in the consecutive rows and columns. For conditions sampled from the small range, all pairs of conditions are compared approximately the same number of times. However, the number of comparisons gradually decreases for conditions further away in the scale, i.e. further away from the diagonal on the heatmap. For conditions sampled from the medium and large ranges, most comparisons are selected for conditions close in the quality scale, i.e. along the diagonal on the heatmap. This is expected, as pairs of conditions that are far away in the scale are less likely to be confused by observers and are therefore less informative.
 
\subsection{Confidence intervals for large range}
\begin{figure}[t]
    \centering
\includegraphics[width=0.95\linewidth]{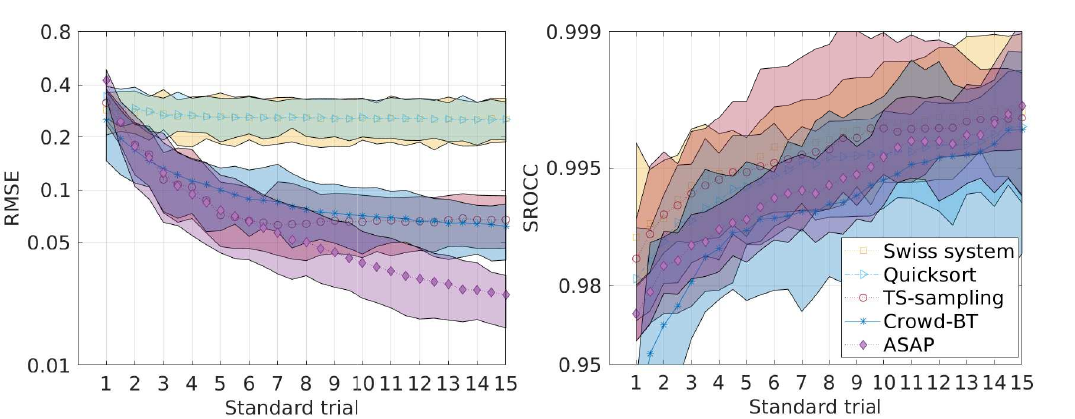}
  \caption{75\% of the RMSE and SROCC distributions for 20 conditions sampled from the large range and five best performing methods in terms of SROCC.}
  \label{fig:ci} 
\end{figure}

Figure \ref{fig:ci} presents 75\% of the RMSE and SROCC distributions for the top five methods in terms of SROCC for conditions sampled from the large range (bottom row of Figure 6 in the main paper). Results for SROCC are noisier than for RMSE, making it hard to identify the best performing method. In terms of RMSE for a number of standard trials less than two ASAP shows similar to others performance, however with the number of standard trials growing, significantly outperforms the compared methods.

\subsection{Results: VQA dataset}
\begin{figure*}[ht!]
    \centering
           
    \subfloat[Reference 3]{%
       \includegraphics[width=0.48\linewidth]{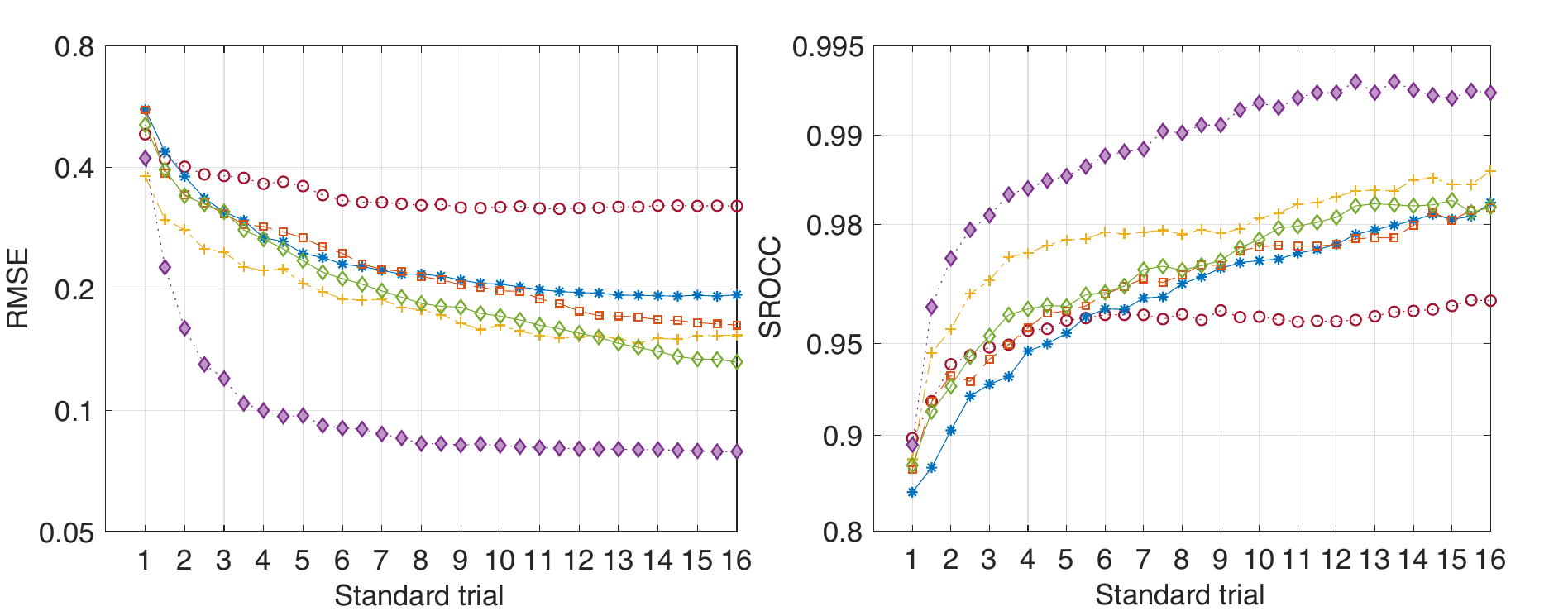}
       }
      \subfloat[Reference 4]{%
       \includegraphics[width=0.48\linewidth]{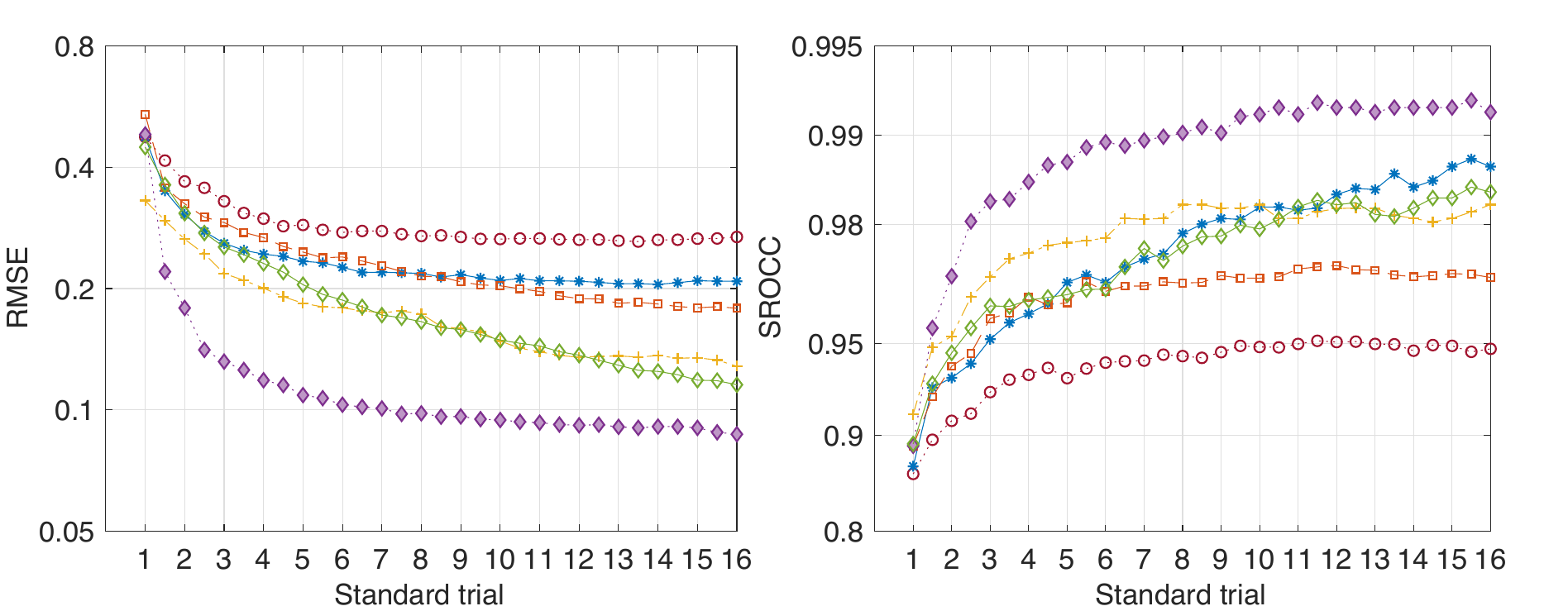}
       }
       
    \subfloat[Reference 5]{%
       \includegraphics[width=0.48\linewidth]{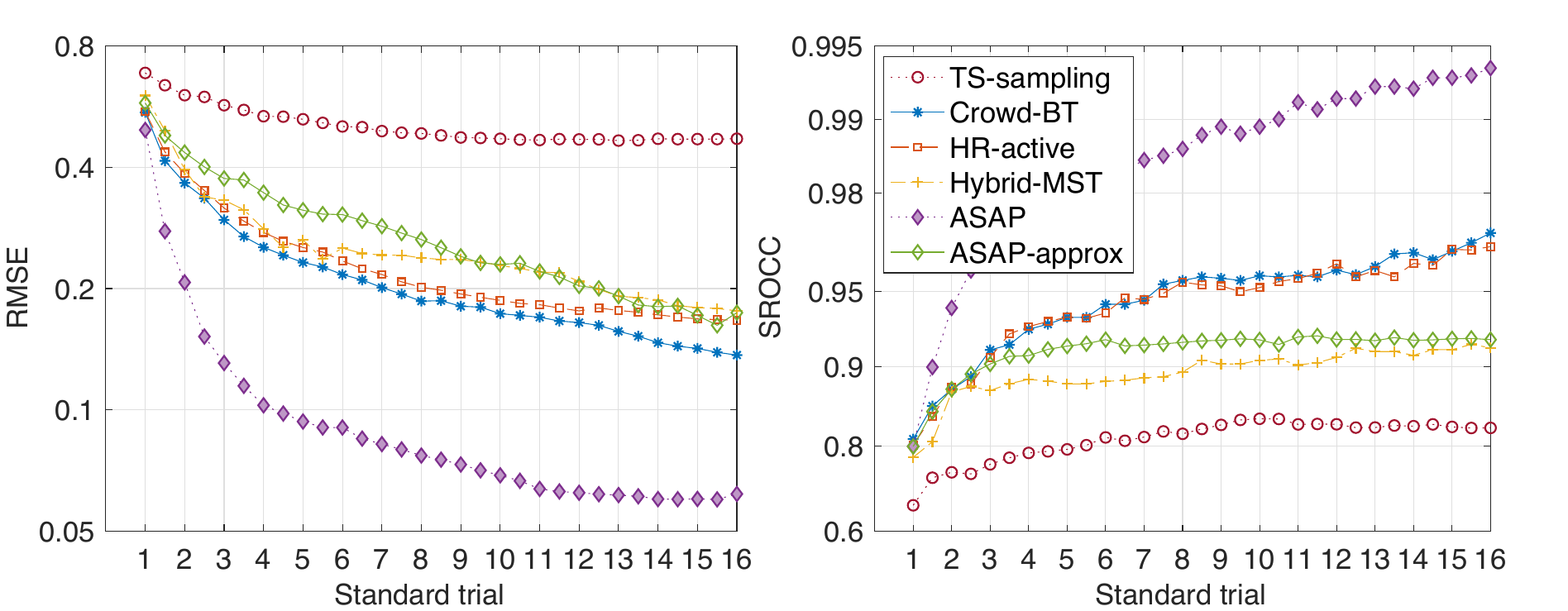}
       }
    \subfloat[Reference 6]{%
       \includegraphics[width=0.48\linewidth]{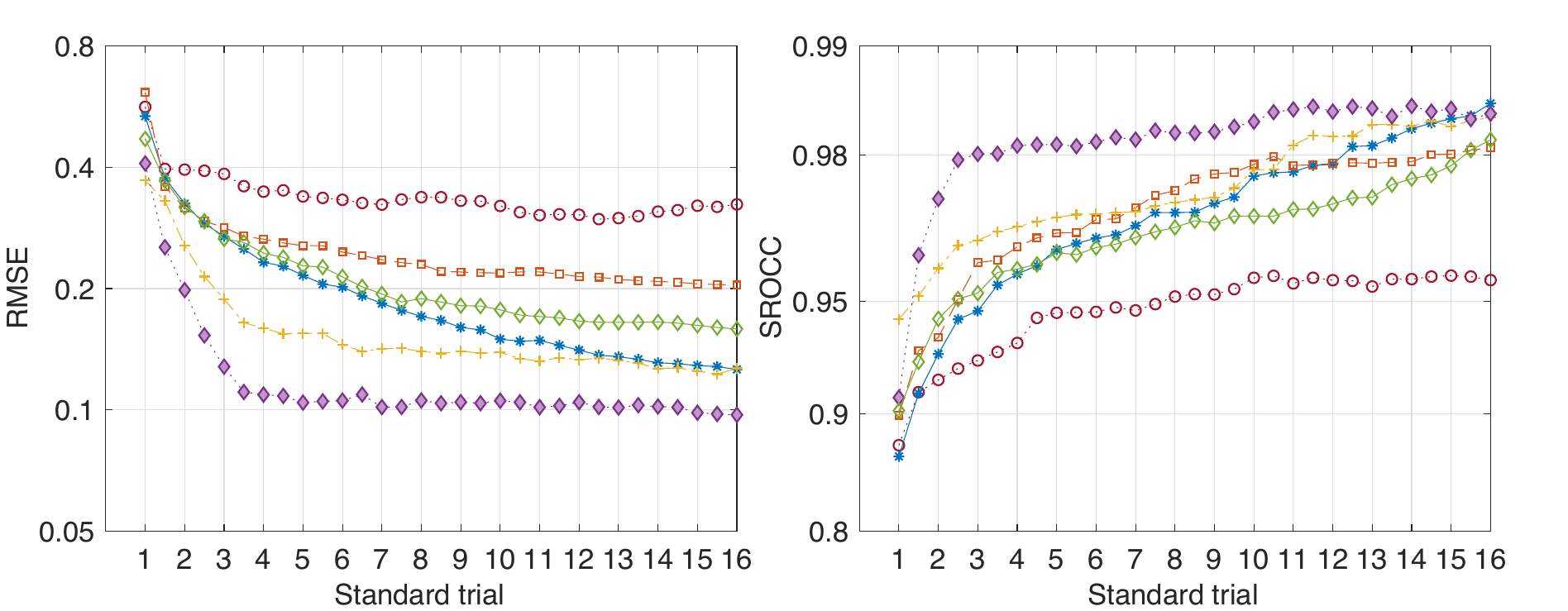}
       }
       
  \subfloat[Reference 7]{%
       \includegraphics[width=0.48\linewidth]{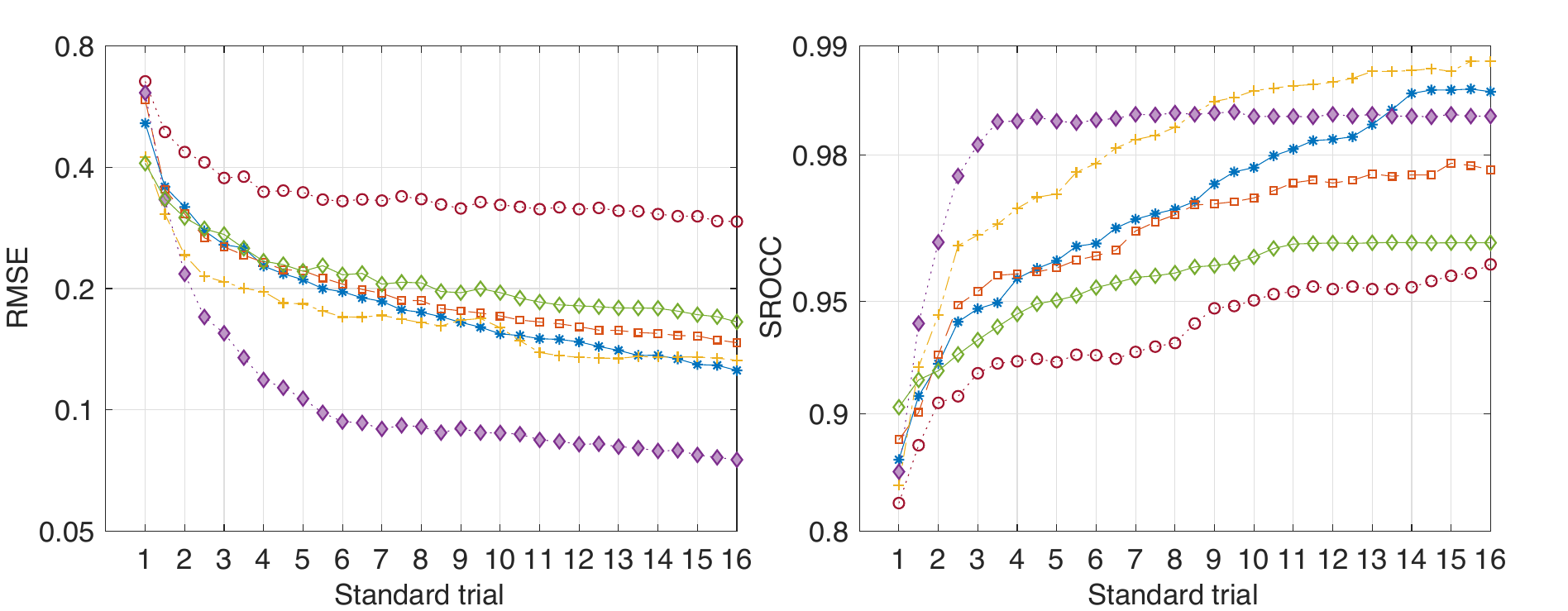}
       }
  \subfloat[Reference 8]{%
       \includegraphics[width=0.48\linewidth]{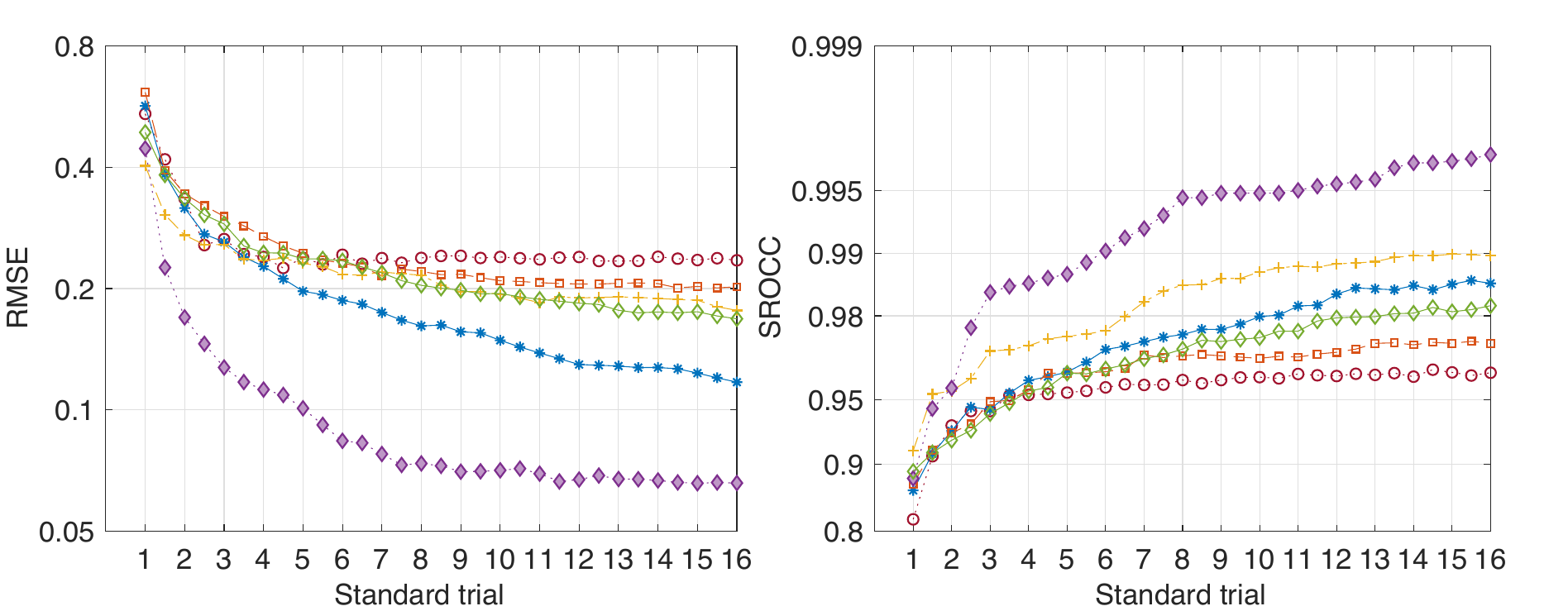}
       }
       
  \subfloat[Reference 9]{%
       \includegraphics[width=0.48\linewidth]{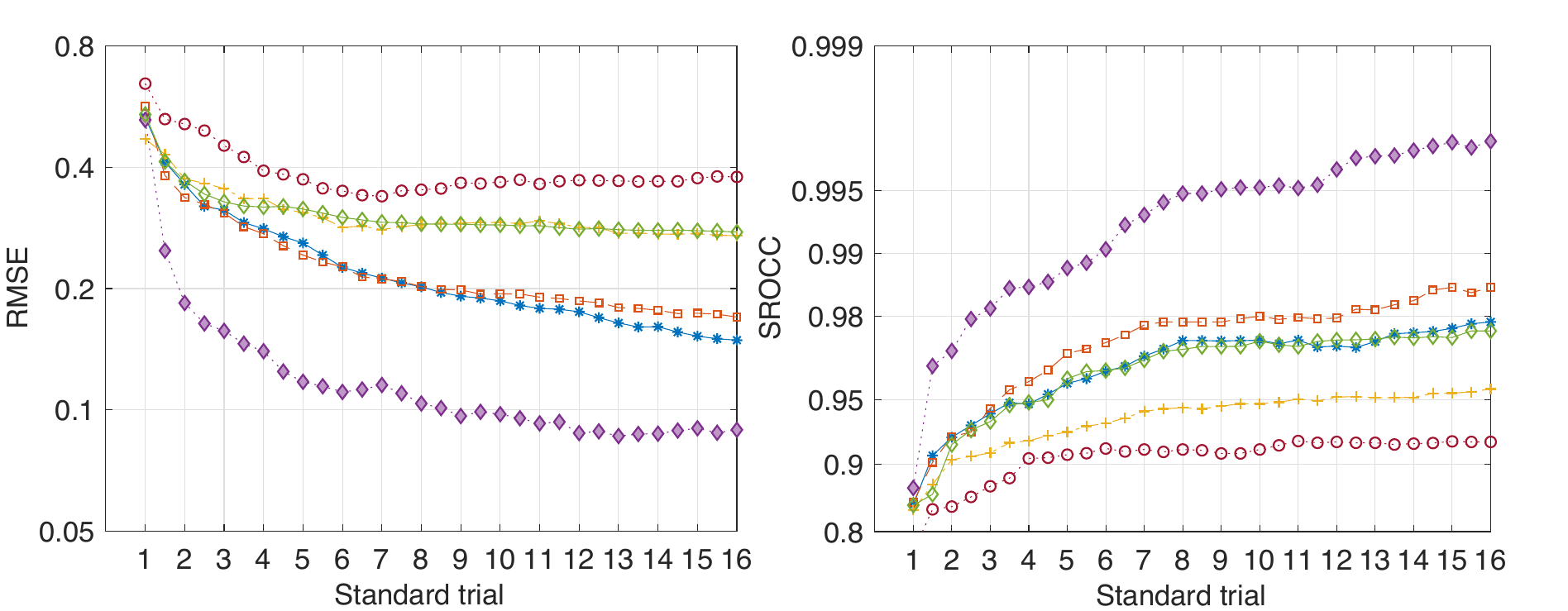}
       }      
  \subfloat[Reference 10]{%
       \includegraphics[width=0.48\linewidth]{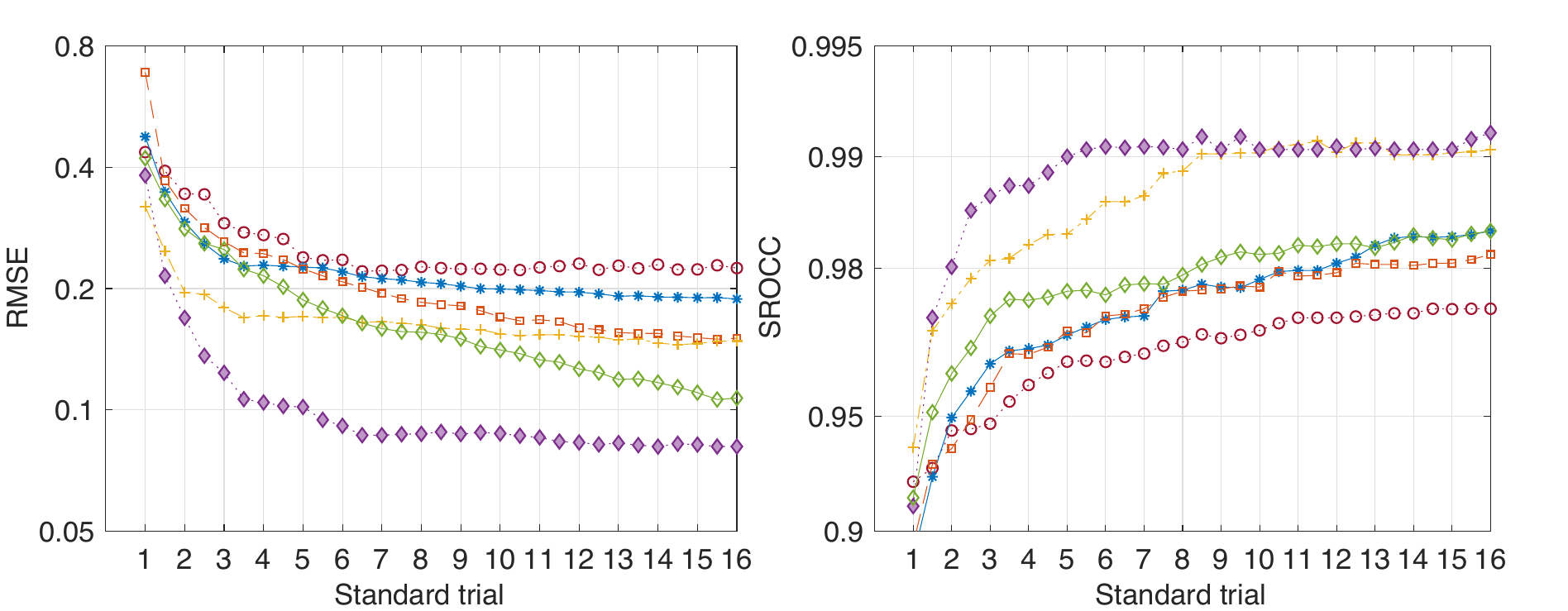}
       }       
  \caption{Compared sampling strategies on VQA dataset.}
  \label{fig:vqa} 
\end{figure*}

This dataset contains 10 reference videos with 16 distortions applied to them. Each $16\times16$ matrix contains 3840 pairwise comparisons - each pair was compared 32 times. Figure \ref{fig:vqa} shows the results for reference videos 3 to 10. Results for the first two reference videos are presented in the main paper. Consistent with other tests in the main paper, ASAP shows superior results to other methods. ASAP-approx. has average, similar to other EIG based methods, results. Hybrid-MST tends to perform better for small numbers of standard trials. 

%\section*{Acknowledgment}




%\bibliographystyle{IEEEtran}
%\bibliography{IEEEabrv,ref}
%\begin{thebibliography}{1}

%H.~Kopka and P.~W. Daly, \emph{A Guide %to \LaTeX}, 3rd~ed.\hskip 1em plus

%\end{thebibliography}
